%% file: main_cdc.tex
\documentclass[letterpaper, 10 pt, conference]{cdc-style-files/ieeeconf}  % Comment this line out
\IEEEoverridecommandlockouts                              % This command is only
\usepackage{microtype}
\usepackage{graphicx}
\usepackage{float}
\usepackage{booktabs}
\usepackage{multirow}
\usepackage{tablefootnote}

\usepackage{caption}
\usepackage{subcaption}

\usepackage{amsmath, amssymb, amsthm, mathtools}

\usepackage{csquotes}

\usepackage{tikz}
\usetikzlibrary{shapes, arrows, positioning, fit, backgrounds, calc, shadows, decorations.pathmorphing}
\definecolor{inputColor}{HTML}{4A90E2}
\definecolor{blockColor}{HTML}{5AC8FA}
\definecolor{smallBlockColor}{HTML}{34C759}
\definecolor{transformerColor}{HTML}{FF9500}
\definecolor{outputColor}{HTML}{FF3B30}
\definecolor{borderColor}{HTML}{1C1C1E}
\definecolor{commentColor}{HTML}{8E8E93}

\usepackage[textsize=tiny]{todonotes}

\usepackage{hyperref}
\usepackage[capitalize,noabbrev]{cleveref}

\theoremstyle{plain}

\theoremstyle{definition}

\theoremstyle{remark}

\title{\LARGE \bf
Flash STU: Fast Spectral Transform Units}

\author{ \parbox{1.5 in}{\centering Y. Isabel Liu*
        \thanks{*Equal contribution. Order determined alphabetically by last name.}\\
        Princeton University\\
        {\tt\small isabel.liu@princeton.edu}}
        \hspace*{ 0.5 in}
        \parbox{1.5 in}{ \centering Windsor Nguyen*\\
        Princeton University\\
        {\tt\small mn4560@princeton.edu}}
        \hspace*{ 0.5 in}
        \parbox{1.5 in}{ \centering Yagiz Devre\\
        Princeton University\\
        {\tt\small yagiz.devre@princeton.edu}}\\
        \\
        \parbox{1.5 in}{\centering Evan Dogariu \\
        New York University\\
        {\tt\small ed2719@nyu.edu}}
        \hspace*{ 0.5 in}
        \parbox{1.5 in}{ \centering Anirudha Majumdar\\
        Princeton University\\
        {\tt\small ani.majumdar@princeton.edu}}
        \hspace*{ 0.5 in}
        \parbox{1.5 in}{ \centering Elad Hazan\\
        Princeton University\\
        {\tt\small ehazan@princeton.edu}}
}

\begin{document}

\maketitle
\thispagestyle{empty}
\pagestyle{empty}

\input{0.abstract}

\input{01_intro}
\input{02_comparisons}

\input{03.LLM}
\input{04_future_work}

\input{06.acks}

\bibliographystyle{ieeetr}
\bibliography{main}

\onecolumn

\include{appendix}
\include{appendix_llm}

\end{document}

%% file: 0.abstract.tex
\begin{abstract}
\noindent
Recent advances in state-space model  architectures have shown great promise for efficient sequence modeling, but challenges remain in balancing computational efficiency with model expressiveness. We propose the \textbf{Flash STU} architecture, a hybrid model that interleaves spectral state space model  layers with sliding window attention, enabling scalability to billions of parameters for language modeling while maintaining a near-linear time complexity. We evaluate the Flash STU and its variants on diverse sequence prediction tasks, including linear dynamical systems, robotics control, and language modeling. We find that, given a fixed parameter budget, the Flash STU architecture consistently outperforms the Transformer and other leading state-space models such as S4 and Mamba-2.

% Our model builds upon the Spectral Transform Unit (STU), a theoretically grounded SSM architecture with a near-linear time complexity that capitalizes on the eigenstructure decay of a special Hankel matrix to efficiently extract principal information in symmetric linear systems.

%These results suggest the STU as a promising primitive for long-range sequence modeling. 

%We release a fully distributed open-source training codebase\footnote{\url{https://github.com/hazan-lab/flash-stu/}} to encourage broader community adoption.  
%%%% TBD after annomymization
\end{abstract}

%% file: 01_intro.tex
\section{Introduction}

Transformers have become the dominant architecture for sequence modeling due to their powerful self-attention mechanism, which allows them to capture complex dependencies across sequences~\cite{vaswani2017attentionneed}. This capability makes them particularly effective for tasks requiring long-range context and recall such as language modeling. However, Transformers suffer from quadratic computational complexity in sequence length, making them inefficient for processing extremely long sequences.

State Space Models (SSMs), in contrast, offer a more computationally efficient alternative by maintaining a fixed-size latent state that does not grow with sequence length~\cite{gu2022efficientlymodelinglongsequences}. This design enables SSMs to scale more efficiently for tasks requiring long-horizon memory while maintaining subquadratic complexity. Recent architectures, such as S4~\cite{gu2022efficientlymodelinglongsequences} and Mamba~\cite{gu2024mambalineartimesequencemodeling}, have demonstrated that SSMs can match or even outperform Transformers in certain structured sequence prediction tasks, particularly in domains like dynamical systems, robotics, audio modeling, and time-series forecasting. However, SSMs face limitations in capturing complex nonlinear dependencies and may struggle with tasks requiring exact retrieval of past inputs, such as copying mechanisms or highly structured symbolic reasoning~\cite{dao2024transformersssmsgeneralizedmodels}.

Our work contributes to this discussion by building on the Spectral Transform Unit~\cite{agarwal2023spectral}, henceforth referred to as the \textit{STU}, a novel state-space-inspired model that integrates spectral filtering techniques~\cite{hazan2017learning} for robust and efficient sequence modeling. Unlike standard SSMs, STU employs fixed convolutional filters that do not need to be learned, derived from a robust theoretical foundation.

We introduce a new \textit{hybrid} model architecture called \textbf{Flash STU}\footnote{Our optimized LLM pretraining code is open sourced at \url{https://github.com/hazan-lab/flash-stu/}.} which combines the STU and sliding window attention~\cite{beltagy2020longformerlongdocumenttransformer}. We observe that Flash STU achieves superior performance over both Transformers and other leading SSM variants on sequence prediction tasks across multiple modalities, including natural language, robotics, and synthetic dynamical systems. We also investigate the optimization landscape of STU in comparison to other models and find that STU models are easier to optimize and provide more stability during training.

\begin{figure}[ht]
    \centering
    \includegraphics[width=1.0\columnwidth]{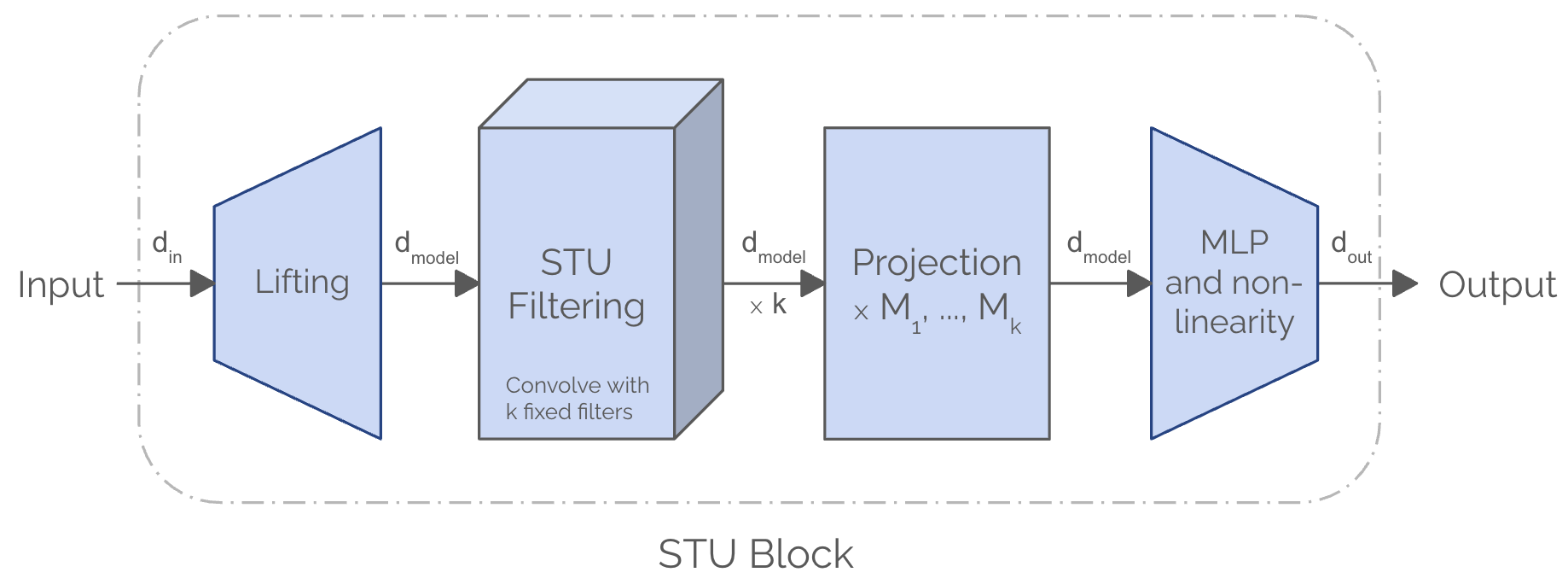}
    \vspace{-1em}
    \caption{Basic architecture of the Spectral Transformer Unit.}
    \label{fig:arch}
\end{figure}

\vspace{-1em}
\subsection{Learning in linear dynamical systems}

Many modern neural architectures for sequence modeling can be understood through the lens of learning in \textit{linear dynamical systems (LDS)}, a fundamental class of models in control theory, time-series forecasting, and machine learning. LDSs are defined by the state-space equations:
\begin{align*}
x_{t+1} & = A x_t + B u_t + w_t \\
y_t & = C x_t + D u_t + \zeta_t
\end{align*}
where \( x_t \in \mathbb{R}^{d_{\text{hidden}}} \) is the hidden state, \( u_t \in \mathbb{R}^{d_{\text{in}}} \) the input, and \( y_t \in \mathbb{R}^{d_{\text{out}}} \) the observed output. Matrices \( A, B, C, D \) parameterize the system, and \( w_t, \zeta_t \) represent noise terms. 

Expanding these recursive equations in the noiseless case (assuming \( D = 0 \)), we obtain:
\begin{equation*}
    y_t = \sum_{i=1}^{t} C A^i B u_{t-i}
\end{equation*}
which explicitly shows how the system’s past inputs contribute to future outputs. The spectral radius \( \rho(A) \), which measures the largest absolute eigenvalue of \( A \), determines how much past information is retained. If \( \rho(A) \approx 1 \), the system has long memory, but this also makes learning difficult due to slow decay in dependencies.

\textbf{Challenges in learning LDS.}
Learning in LDS involves predicting future observations given past inputs, which becomes particularly challenging when the system is \textit{marginally stable} (i.e., \( \rho(A) \approx 1 \)) due to long memory effects. Classical approaches face significant limitations: System Identification estimates A, B, C, D directly but is non-convex and unreliable for marginally stable systems; Autoregressive Learning requires many parameters for long-memory systems~\cite{hazan2022introduction}; and Kalman Filtering is sensitive to adversarial settings and instability ~\cite{kalman1960new,anderson1991kalman}.

A more recent perspective on learning in LDS involves \textit{spectral filtering} \cite{hazan2017learning,hazan2018spectral}, which focuses on spectral properties of the transition matrix \( A \) rather than direct estimation of system parameters. The core idea is to represent sequence dynamics using \textit{fixed convolutional filters} derived from dominant spectral components of the system.

This approach has several advantages:
\begin{enumerate}
    \item \textbf{Bypasses direct parameter estimation} – Unlike system identification, spectral filtering does not require recovering \( A, B, C, D \), avoiding issues with non-convex optimization. Further, the theoretical properties of spectral filtering suggests that the number of parameters do not scale with the hidden dimension of the system, or the dimension of the transition matrix $A$. 
    
    \item \textbf{Handles marginal stability} – Spectral filtering is capable to learn marginal stable systems with symmetric transition matrices, namely: it remains effective even when \( \rho(A) \approx 1 \). This has the intuitive capability of learning in the presence of long context, a theoretical advantage that we verify experimentally henceforth. 
    
    \item \textbf{Computational efficiency} – Instead of maintaining a full latent state or learning a long Markov operator, spectral filtering converts the input sequence to a compressed form. As a convolutional model, it can be implemented using the fast Fourier transform algorithm for computing convolutions, enabling subquadratic complexity.
\end{enumerate}

The \textit{Spectral Transform Unit (STU)} extends spectral filtering techniques to sequence modeling. Instead of learning full transition matrices, STUs apply fixed spectral filters derived from the dominant eigenvectors of a Hankel matrix, ensuring that past information is captured efficiently. This enables stable long-range sequence modeling with significantly lower computational cost compared to Transformers and traditional state-space models.

%In the next section, we detail the STU architecture and its advantages for sequence prediction.

\subsection{Description of the Spectral Transform Unit}
The STU architecture is depicted in Figure~\ref{fig:arch}. Given an input sequence, STU transforms it through the following steps: first, the input may be lifted to a higher-dimensional space via a learned transformation. Then, a convolution operation is applied using a set of \textbf{fixed} filters that do not require training, followed by a learned transformation that projects the filtered sequence to the output space. Finally, an optional nonlinearity is applied.

Mathematically, the output is given by:
\begin{equation}
    \hat{y}_t = \sigma \left( \sum_{i=1}^k M_i \cdot \langle \Phi_i , u_{t:t-L} \rangle \right),
\end{equation}
where \( M_i \) are learned projections, \( \sigma \) is a nonlinearity, and \( \Phi_{1:k} \) are fixed convolutional filters that can be precomputed. The filters \( \Phi_{1:k} \) correspond to the eigenvectors of the largest eigenvalues of the Hankel matrix:
\begin{equation}
    Z = \int_{\alpha=0}^1 \mu_\alpha \mu_\alpha^\top d\alpha, \quad \mu_\alpha = (1, \alpha, ..., \alpha^L).
\end{equation}
These filters can be computed \textbf{a priori} and stored before observing the actual sequence, making STU particularly efficient.

The mathematical properties that underline spectral filtering are derived from the following remarkable fact about Hankel matrices: the spectrum of any Hankel matrix over the real numbers decays exponentially~\cite{beckermann2017singular}. This fact was used by~\cite{hazan2017learning} to design a filtering basis that is very sparse in terms of the number of filters needed. The residual error, due to this Hankel property, is exponentially small. 
More details about the spectral filters are described in Appendix \ref{appendix:spectral-filters}.

%\subsection{Tensordot approximation (STU-T)}
A key optimization technique is the tensordot approximation (\textbf{STU-T}) \cite{agarwal2023spectral}, which decomposes the projection tensors \( M_i \) into two smaller matrices:
\begin{equation}
    M_i \approx M_i^1 \times M_i^2, \, M_i^1 \in \mathbb{R}^{d_{\text{model}} \times k}, \, M_i^2 \in \mathbb{R}^{d_{\text{model}} \times d_{\text{out}}}
\end{equation}
This approximation reduces computational complexity by approximately a factor of \( k \), saving memory and leading to significant improvements in efficiency. Although the expressivity of the model is slightly reduced under this regime, we show that STU-T retains competitive empirical performance, making it a practical alternative for scaling spectral state-space models efficiently.

%% file: 02_comparisons.tex
\section{Experiments with synthetic data}
\label{sec:LDS experiments}
We begin our investigation of STU's properties with some simple yet representative synthetic tasks that have become commonplace in the sequence modeling literature. In particular, we aim to understand the behavior of STU in environments with long memory and nonlinearities, especially as we introduce feed-forward layers and deeper architectures. We compare against S4~\cite{gu2022efficientlymodelinglongsequences} and Mamba-2~\cite{dao2024transformersssmsgeneralizedmodels}, standard SSM architectures, as well as the vanilla transformer layer~\cite{vaswani2017attentionneed}. Note that Mamba-2 contains a nonlinear selection mechanism that can be beneficial in certain synthetic tasks at the cost of extra complexity and the loss of the convolutional inference mode. Nonlinear gating can also be added to STU layers for use in larger models, as done in Section \ref{robotics-experiments}.

% \paragraph{Linear Dynamical System:} 
\subsection{Linear dynamical systems}
%In a linear dynamical system (LDS), the outputs $\{y_t\}$ of a sequence are generated according to the dynamics equations \[ x_{t} = A x_{t-1} + B u_t, \ y_{t} = C x_t + D u_t, \] where $\{u_t\}$ is an (observed) input sequence and $\{x_t\}$ are hidden states of the system. The matrices $A,B,C, D$ are called "system matrices" and are unknown to the sequence predictor. Typical approaches to learning in this setting scale in complexity with hidden state dimension $d_{\operatorname{hidden}}$ and effective system memory $1/\delta$, where $\rho(A) = 1-\delta$ is the spectral radius of $A$. For an in-depth treatment of linear dynamical systems, as well as methods to learn and predict them, see the text~\cite{hazan2022introduction}. \\

\begin{figure}[h]
    \centering
    \begin{minipage}{0.7\columnwidth}
        \centering
        \includegraphics[width=\textwidth]{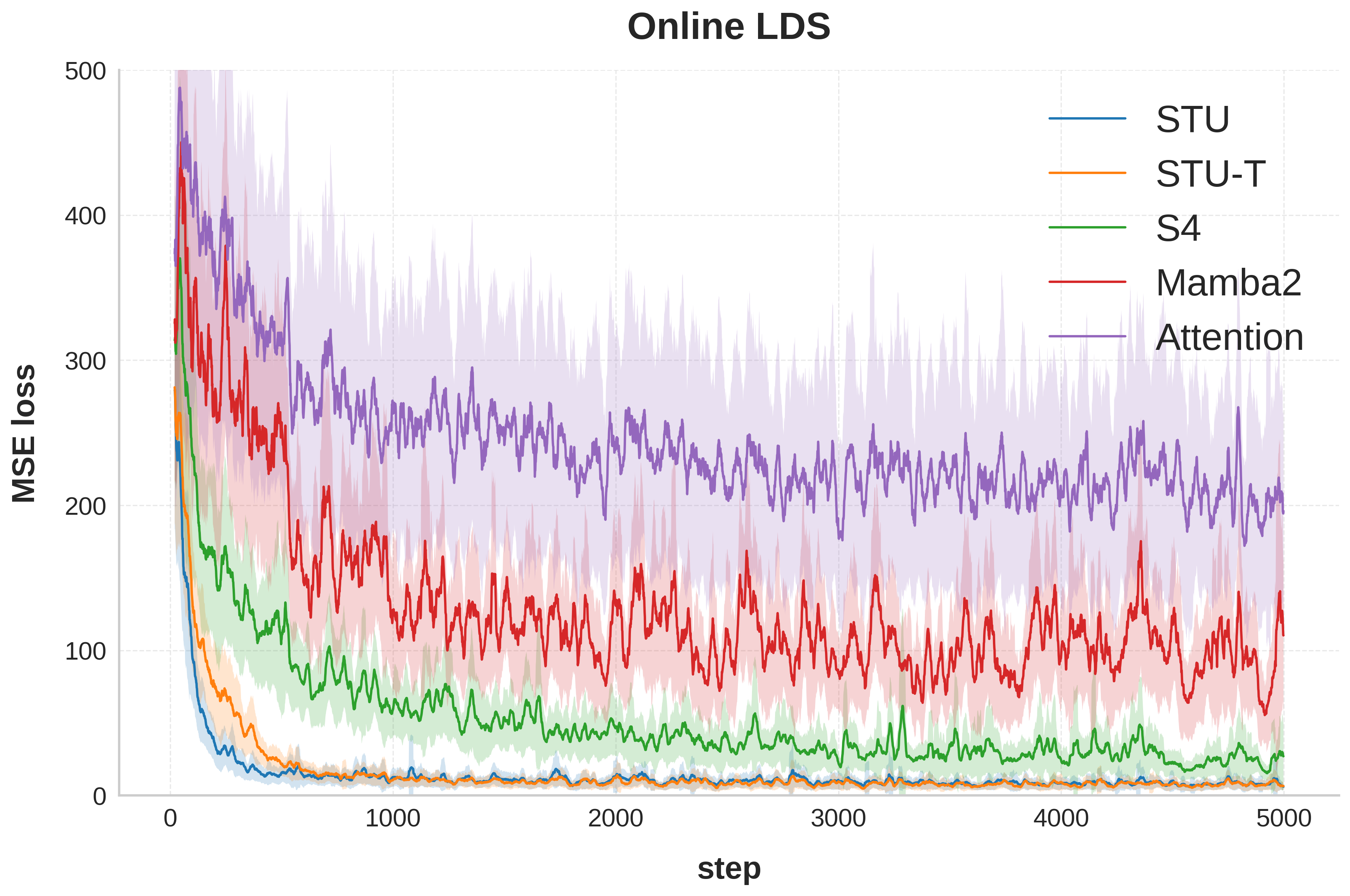}
        \caption{Mean squared error $\|\hat{y}_{t+1} - y_{t+1}\|^2$ of the different layers on a single sequence from an LDS.}
        \label{fig:synthetics-ldsplot}
    \end{minipage}
\end{figure}

\textbf{Experiment details and conclusions.} We apply random inputs to linear dynamical systems with random system matrices with $d_{\operatorname{input}}=d_{\operatorname{output}}=5$ and hidden state dimension $d_{\operatorname{hidden}} = 256$.
The transition matrix $A$ is symmetrized and normalized so that $\rho(A) = 0.99$ to give the system an effective memory of $\approx 100$, which is what we set the context length to. Such a setting has the properties of (1) long memory and (2) a large hidden state, which we believe to be major sources of complexity in real-world applications.

We compare against a standard attention layer with 8 heads using RoPE (Rotary Position Embeddings)~\cite{su2024roformer} and FlashAttention~\cite{dao2022flashattentionfastmemoryefficientexact}, a diagonal S4 layer~\cite{gu2022efficientlymodelinglongsequences}, and a Mamba-2 layer~\cite{dao2024transformersssmsgeneralizedmodels}. All layers have a width of 32 and are trained for 5,000 steps with the RMSProp optimizer. Results are plotted in Figure \ref{fig:synthetics-ldsplot} with error bars over 16 trials.

We can see that the STU layer is a powerful and robust SSM primitive. Both the vanilla STU and its approximate version STU-T are able to reliably achieve small loss in this setting with quick and robust convergence, while the performance of the other methods vary across random seeds. Note that the width of the STU layer does not need to be as large as $d_{\operatorname{hidden}}$ to perfectly capture the dynamics, which is consistent with theory. Furthermore, STU-T approximation is able to roughly match vanilla STU's performance, even on these multi-input multi-output (MIMO) systems.

\subsection{Optimization behavior}
We saw in the linear dynamical system experiment that the STU layers seem to have a comparatively easier time optimizing. This is expected since an STU layer under MSE loss is a convex parameterization, whereas the losses of the other models are non-convex\footnote{For S4, Mamba-2, and any other model that parameterizes an LDS through its system matrices, this non-convexity even grows with sequence length. One can often relax the problem to a convex one by directly learning a convolutional kernel, though at the cost of maintaining parameter count that grows with sequence length. STU is able to avoid both of these issues with a convex and efficient parameterization.}. Following~\cite{li2017visualizinglosslandscapeneural}, we choose two random directions in the high-dimensional space, move along these directions by varying amounts \texttt{x\_step} and \texttt{y\_step}, compute the loss of the model with perturbed parameters for each coordinate pair in these directions, and plot the loss values as heights on a 2D grid.

\begin{figure}[H]
    \centering
    \begin{minipage}{0.35\linewidth}
        \centering
        \includegraphics[width=\linewidth]{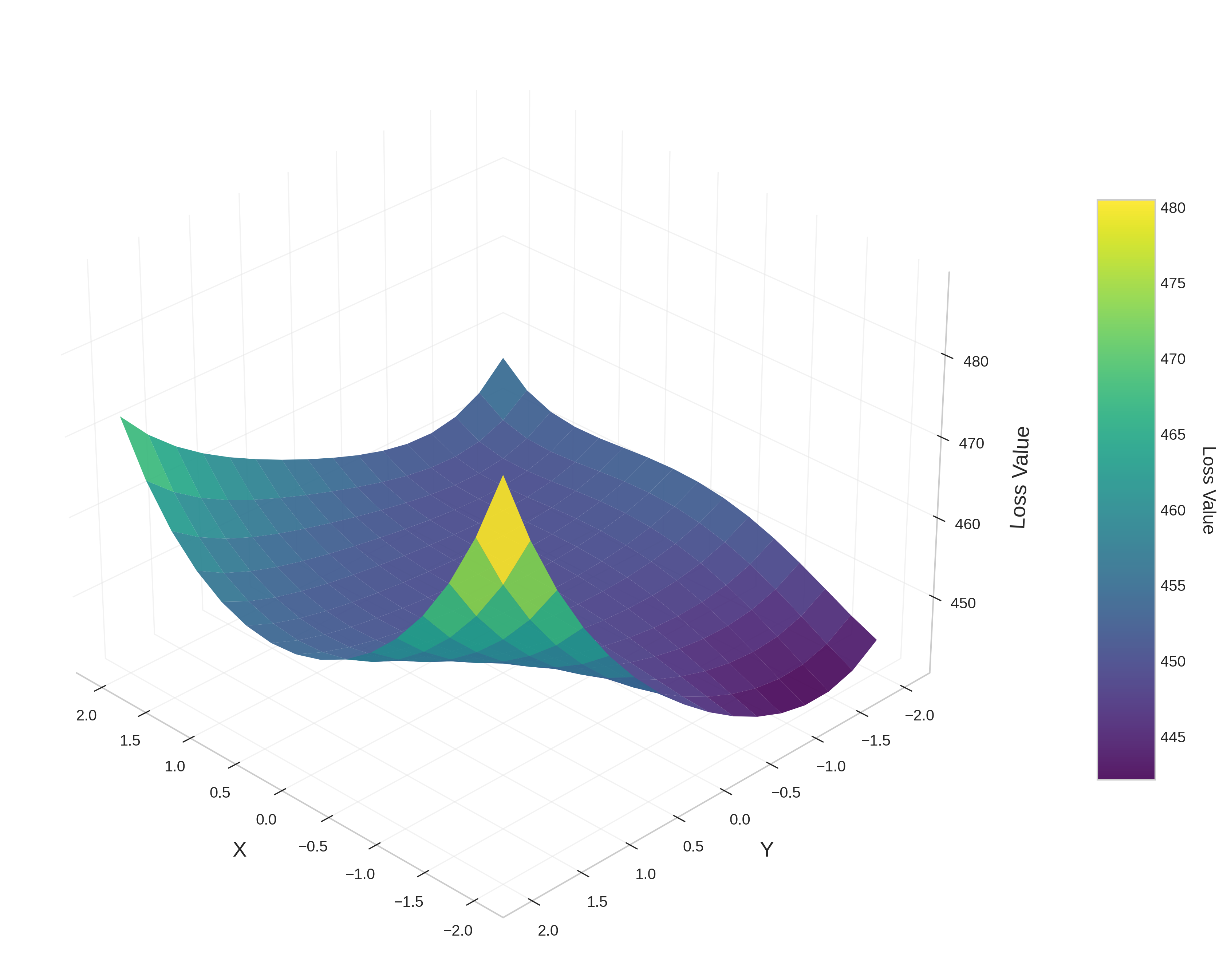}
    \end{minipage}
    \hspace{0.05\linewidth}
    \begin{minipage}{0.35\linewidth}
        \centering
        \includegraphics[width=\linewidth]{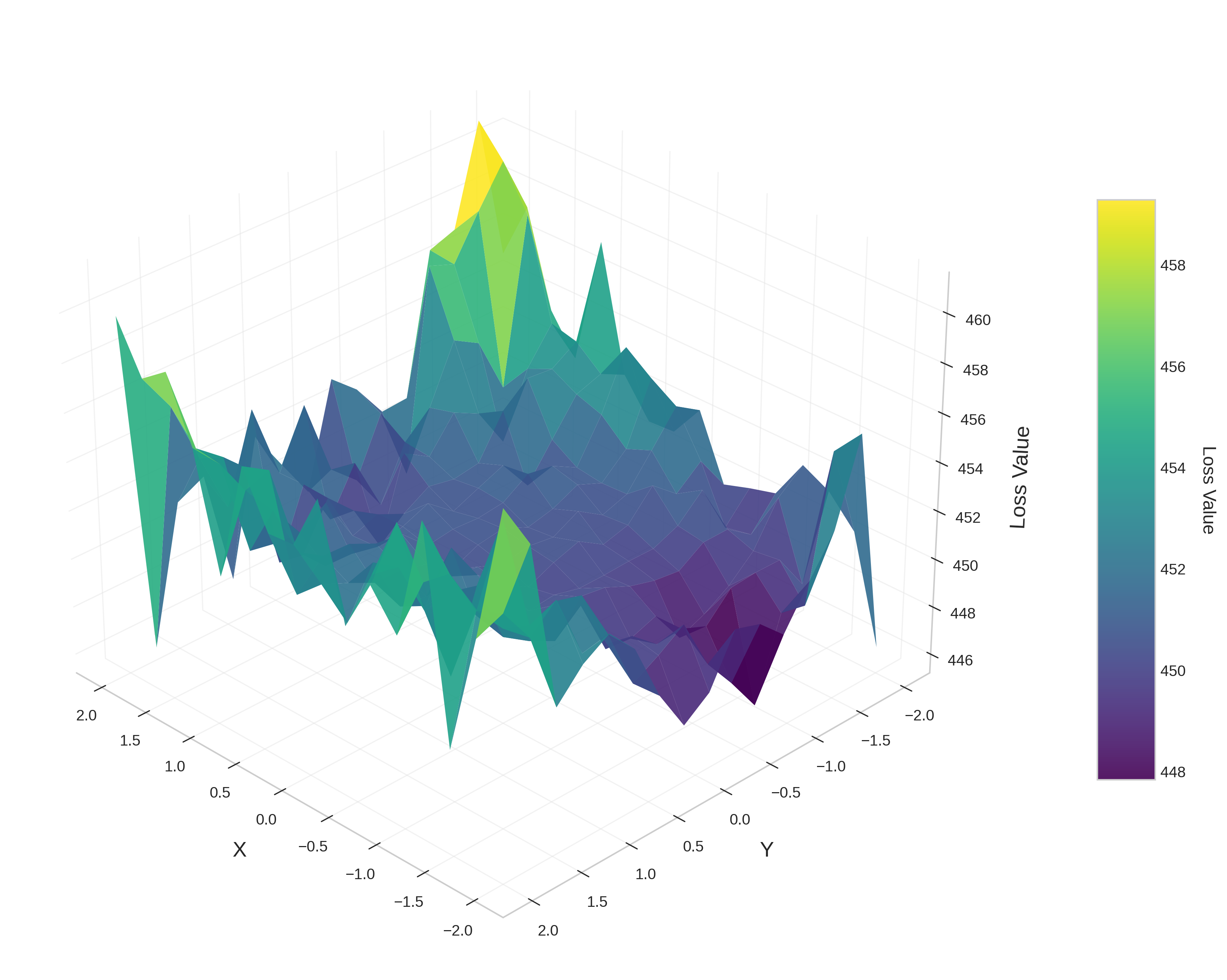}
    \end{minipage}
    \begin{minipage}{0.35\linewidth}
        \centering
        \includegraphics[width=\linewidth]{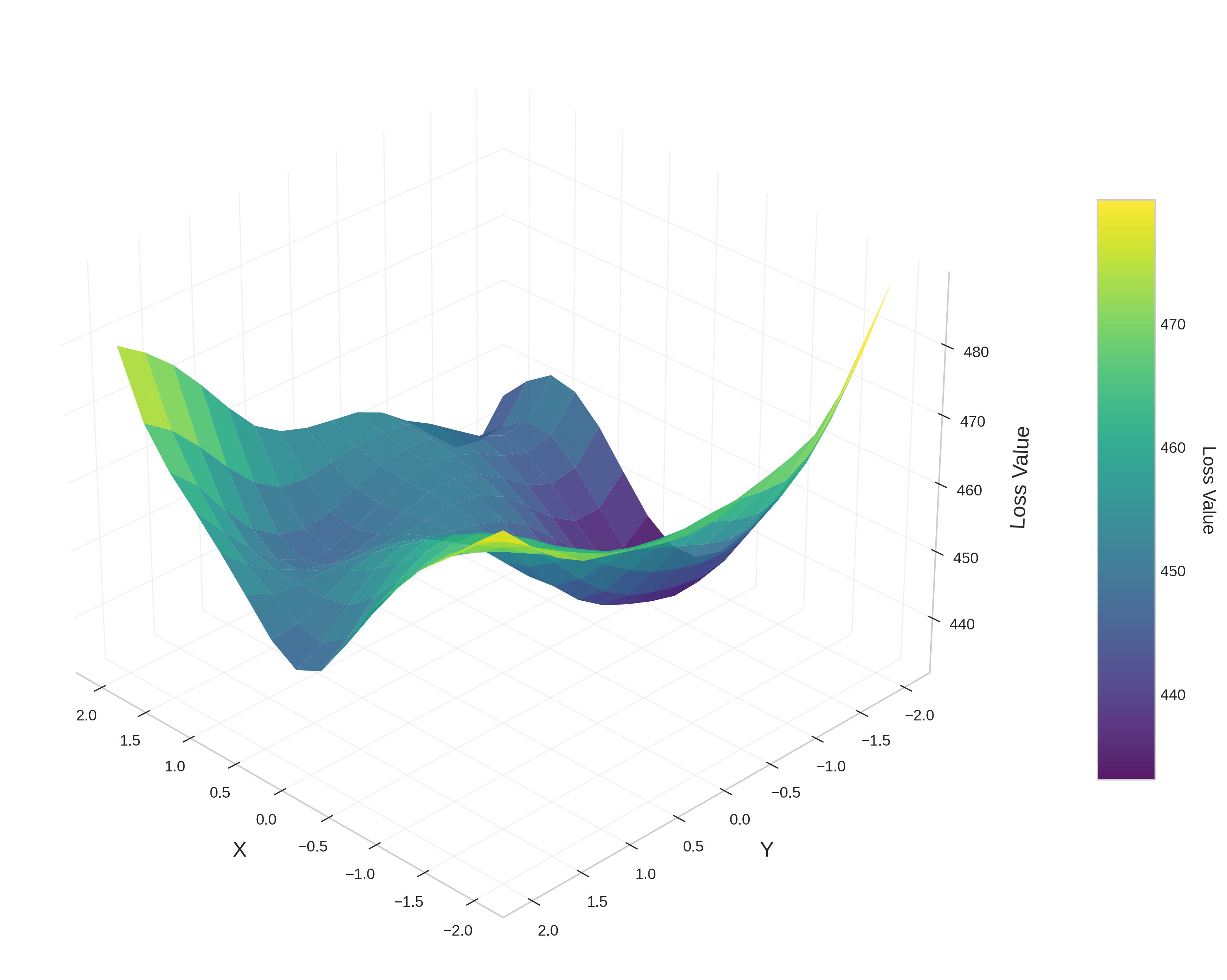}
    \end{minipage}
    \hspace{0.05\linewidth}
    \begin{minipage}{0.35\linewidth}
        \centering
        \includegraphics[width=\linewidth]{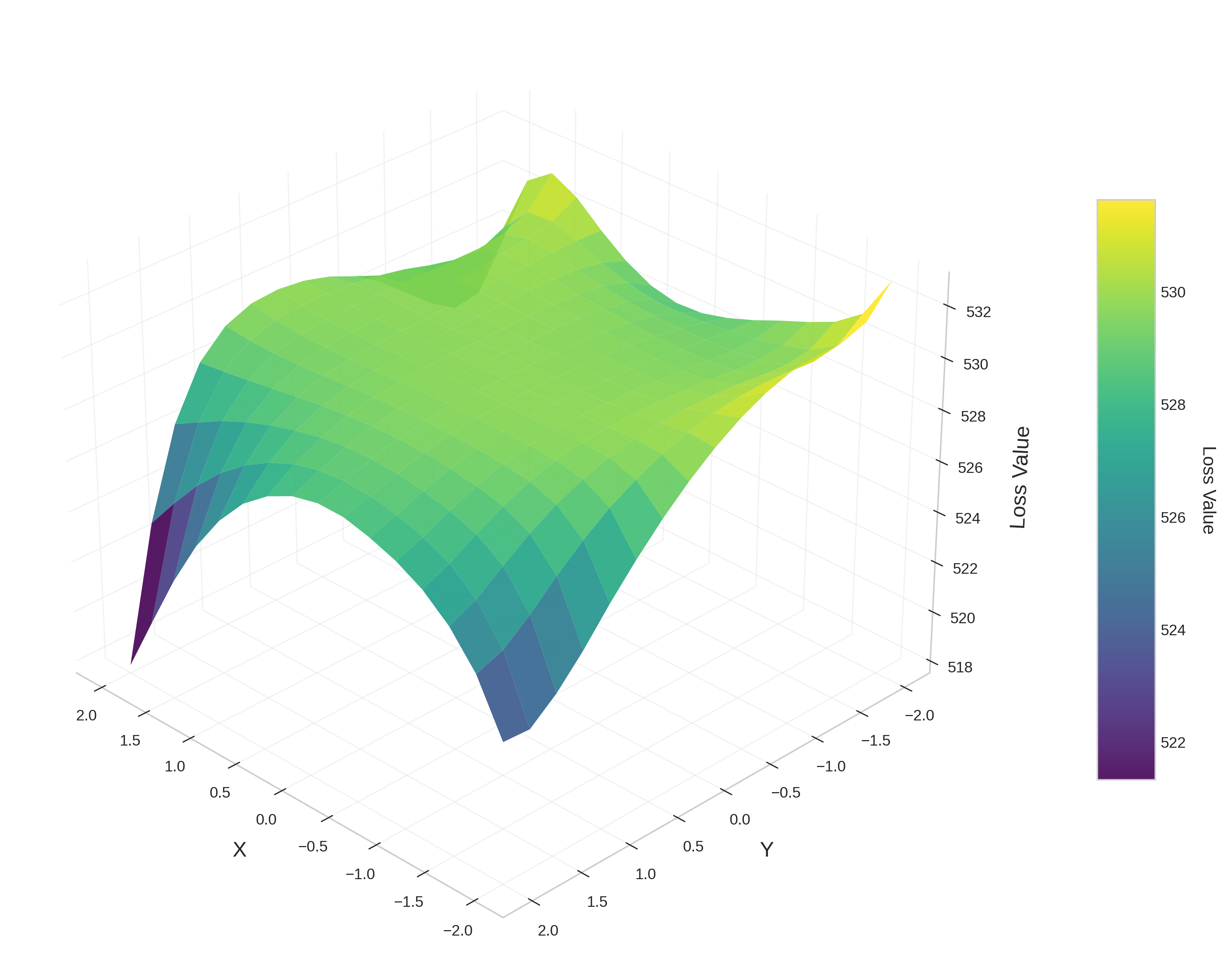}
    \end{minipage}

    \caption{Local loss landscapes for STU, S4, Mamba-2, and attention (in order).}
    \label{fig:synthetics-losslandscapes}
\end{figure}

Figures \ref{fig:synthetics-losslandscapes} visualize these loss landscapes for STU, S4, Mamba-2, and attention layers, respectively, after 10 steps of training on the LDS. Figure~\ref{fig:synthetics-eigvals-heatmaps} in Appendix \ref{appendix:synthetics} instead visualize the local geometry through curvature in terms of Hessian eigenvalue ratios. 
Flatter minima are preferable since it has been proposed that reducing sharpness helps with generalization~\cite{foret2020sharpnessawareminimizationefficientlyimproving}. 
One of the main strengths of STU is its clean optimization landscape. This benefit becomes more important in larger and more complex models.

\vspace{-1em}
\subsection{Other synthetic tasks}
We also investigate the performance of the STU layer in synthetic tasks that are more well-known in the deep learning literature. In the induction heads task, the model is required to recall one token (sampled uniformly from a vocabulary) immediately after a special \verb|flag| token; the rest of the context consists of the same special \verb|blank| token, which the model should learn to ignore. The associative recall task is harder, as it first gives the model an (unordered) sequence of key-value pairs and then asks for the value associated with a query key after. The model must keep track of the entire context, not just a single token. Both of these tasks have nonlinear dynamics and require single-token precision to fully solve. Furthermore, since we apply deeper models, the optimization is non-convex for all the models considered. 

\begin{figure}[H]
    \centering
    \includegraphics[width=0.75\columnwidth]{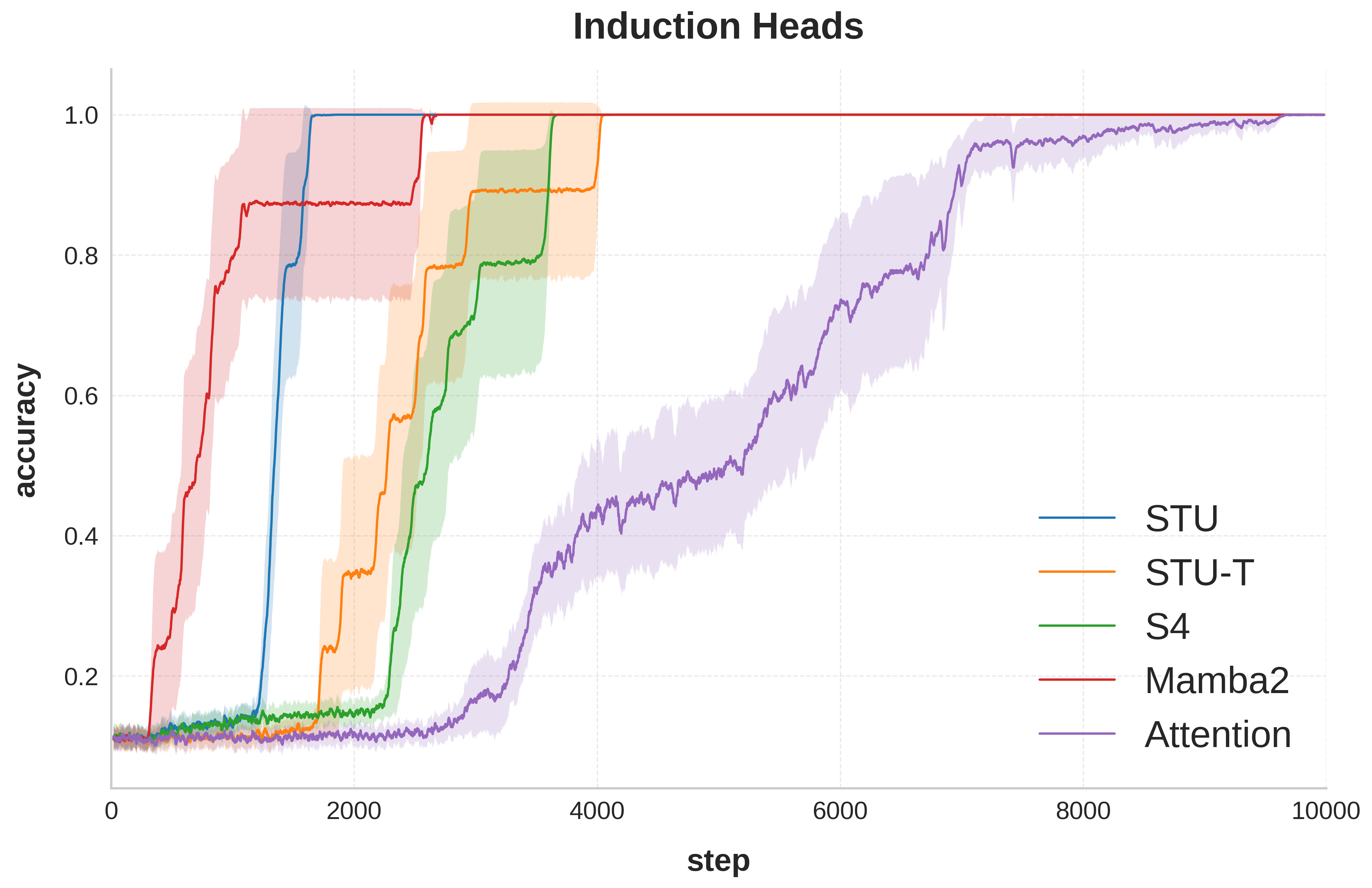}
    \caption{Prediction accuracy for the token immediately following the special flag token during training.
    % \evan{context len = 32, batch size = 128, num steps = 10000, vocab size = 5, depth=2 with mlp, opt = standard adam 1e-3}.
    }
    \label{fig:synthetics-inductionheads}
\end{figure}

\begin{figure}[H]
    \centering
    \includegraphics[width=0.75\columnwidth]{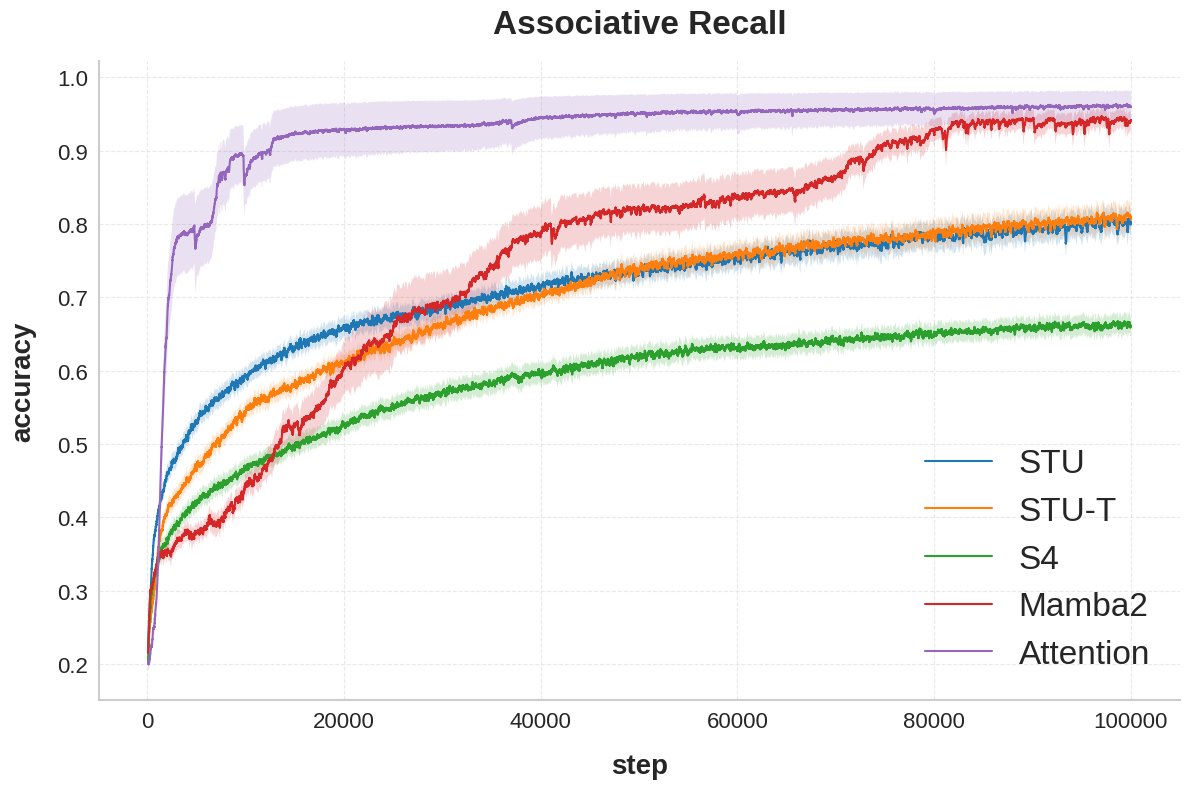}
    \caption{Prediction accuracy for the value corresponding to the given query during training.}
    \label{fig:synthetics-associativerecall}
\end{figure}

\textbf{Experiment details and conclusions.}
We train two-layer models, with MLP layers in between, using the cross-entropy loss, the Adam optimizer with tuned learning rate, and training examples in batches of size 64. For induction heads we set the context length to 128, vocabulary size to 10, and all the model widths to 32. Associative recall was ran with context length 32, vocabulary size of 5, and model widths of 8. Accuracies, averaged over 8 trials, are shown in Figures \ref{fig:synthetics-inductionheads} and \ref{fig:synthetics-associativerecall}.

\noindent
Mamba-2 performs best overall on selection tasks, consistent with its specialized selection mechanism. Attention excels on associative recall, where its ability to retrieve associations in parallel is especially well-suited. Surprisingly, however, STU and STU-T consistently outperform S4 (and even attention on induction heads) with no additional nonlinearities. On induction heads, STU finds solutions more reliably across seeds, suggesting easier optimization. And on the harder associative recall task, it still improves over S4 despite both being convolutional. These results highlight STU as a simple, expressive, and easily optimizable layer for larger models.

% \\ For results on other synthetic tasks, see Appendix \ref{appendix:synthetics}.
% that attention-based methods still seem best-suited for the highly nonlinear, non-dynamical nature of the task. We find it surprising that STU-based models nearly match this, and leave it as future work to understand why some SSM models would perform better than others on this task\footnote{One hypothesis is that the STU layer is able to effectively implement the identity (since the top eigenvector of the Hankel matrix approximates a delta), while a general linear dynamical system cannot; STU is a relaxed model class and is strictly more expressive than a symmetric LDS. We also expect that the convexity of the optimization plays a role -- while these are deep models with MLP's and so are technically non-convex, a recurring theme is that STU appears more well-behaved in terms of optimization dynamics.}. See Appendix \ref{appendix:synthetics} for more details and discussion, as well as results on other related synthetic tasks.

\input{02.2_robotics}

%% file: 02.2_robotics.tex
\section{Experiments with robotics sequence prediction}
\label{robotics-experiments}
We now consider next-state prediction for actuator coordinates using the MuJoCo physics engine~\cite{todorov2012mujoco}, a more challenging sequence prediction task. The goal is to learn the dynamics of a certain physical simulation agent, for example the \texttt{Ant-v1} system,
\begin{align*}
    x_{t+1} = f(x_t, u_t),
\end{align*}
where $x_t \in \mathbb{R}^{29}$ and $u_t \in \mathbb{R}^{8}$ correspond to the state and action at time $t$, respectively.

More precisely, $x_t$ corresponds to the positions, orientations, joint angles, and velocities of the various limbs of the \texttt{Ant-v1} controller, whereas $u_t$ represents the torques applied to the joints of the \texttt{Ant-v1} controller and is generated by an agent pretrained using proximal policy optimization~\cite{schulman2017ppo}. Unlike the synthetic linear dynamical system in Section~\ref{sec:LDS experiments}, the dynamics $f$ for this particular MuJoCo prediction task are nonlinear and hybrid, i.e. non-smooth.

In the experiments below, we defined the loss function to be the squared Euclidean distance between the predicted state vector $\hat{x}_{t+1}$ and the true state vector $x_{t+1}$
\begin{equation*}
L(\theta) = \frac{1}{2} \left\| \hat{x}_{t+1}(\theta) - x_{t+1} \right\|_2^2 = \frac{1}{2} \sum_{i=1}^{n} \left( \hat{x}_{t+1,i}(\theta) - x_{t+1,i} \right)^2 , 
\end{equation*}
where $\hat{x}_{t+1}(\theta)$ is the predicted state vector parameterized by $\theta$, and $x_{t+1}$ is the true state vector.

\textbf{Model architectures.}
We evaluate STU against S4, Transformers, and Mamba-2, with the same 0.5M parameter count for fair comparison. We ensured that they have the same widths and depths and only adjusted the MLP intermediate hidden dimension size as the primary method to equalize the parameter count as commonly seen in the literature~\cite{waleffe2406empirical}. As for the STU-T model, its tensordot approximation allowed for extra parameter space that can be allocated towards other areas in the model. For example, using exactly the same configurations as STU, STU-T only has 0.05M parameters; to scale it up for fair model comparisons, here we choose to expand the ``width" of STU-T, which is the model dimension $d_{\text{model}}$ in Figure \ref{fig:arch}. We also  consider variants augmented with a Mixture-of-Experts (MoE) layer applied after the main sublayer. Each expert is a small gated MLP, and a learned routing mechanism selects the top-$k$ experts per input. 
Figure~\ref{fig:model-architectures} shows the model architectures in detail. 
Table \ref{table:model-specs} gives the specification for each tested architecture in these two sets of experiments.

\begin{figure}[H]
    \centering
    \begin{minipage}{0.58\columnwidth}
        \centering
        \includegraphics[width=\columnwidth]{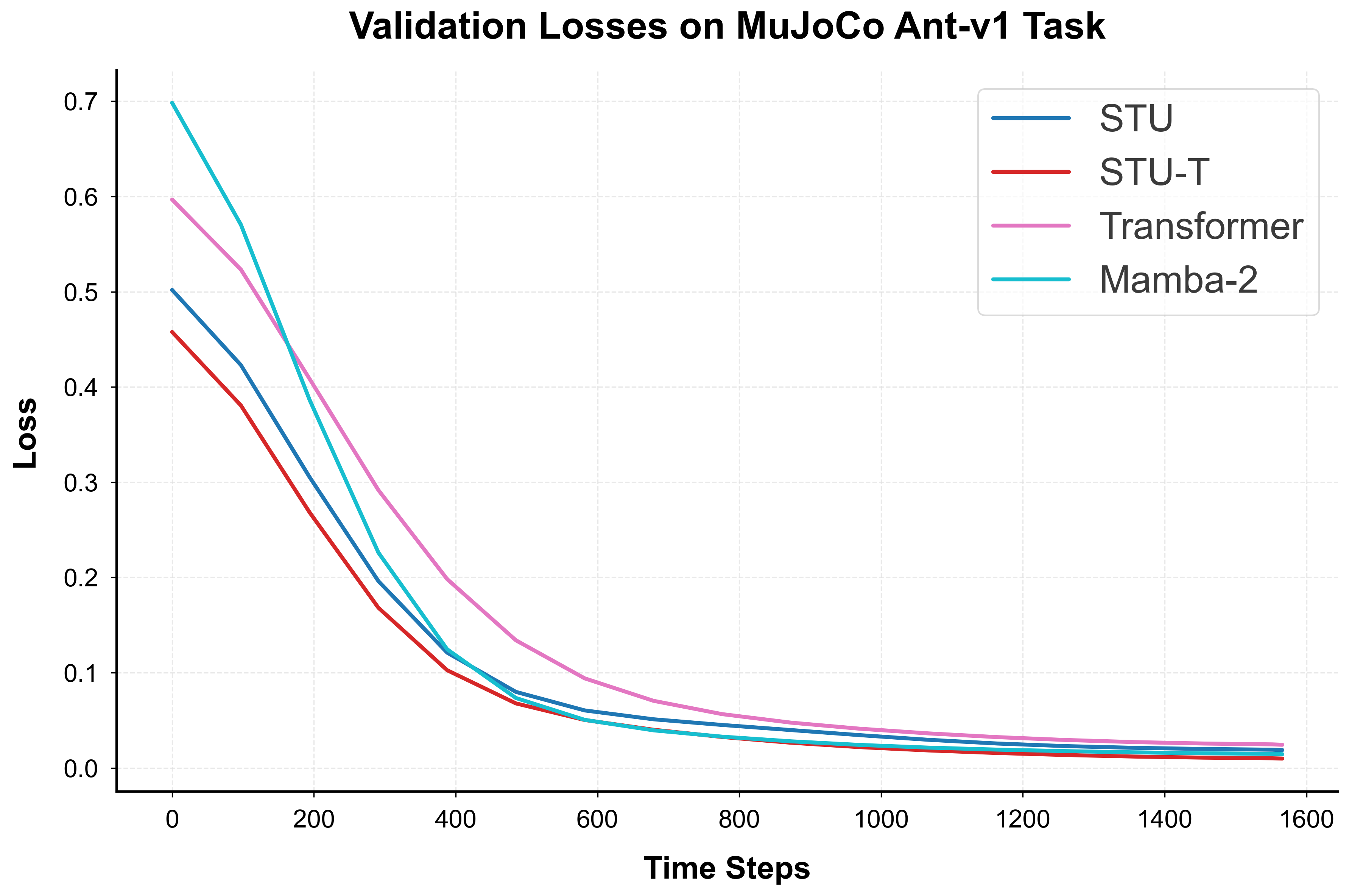}
        \caption{\texttt{Ant-v1} comparative training results.}
        \label{fig:val-losses-Ant}
    \end{minipage}%
    \hfill
    \begin{minipage}{0.38\columnwidth}
        \centering
        \footnotesize
        \setlength{\tabcolsep}{2pt}
        \captionof{table}{\texttt{Ant-v1} comparative validation loss results.}
        \begin{tabular}{@{}lc@{}}
          \toprule
          Model & \begin{tabular}[c]{@{}c@{}}Validation\\Loss\end{tabular}\\
          \midrule
          STU & 0.0181 \\
          STU-T & \textbf{0.0092} \\
          Transformer & 0.0237 \\
          Mamba-2 & 0.0139 \\
          \bottomrule
          \label{table:val-losses-Ant}
        \end{tabular}
    \end{minipage}
\end{figure}

\textbf{Experiment controllers.} We tested our models on three MuJoCo controllers: \texttt{Ant-v1}, \texttt{HalfCheetah-v1}, and \texttt{Walker2D-v1}~\cite{huggingface_ppo_seals}. For each task, we ran a pretrained PPO model over 1,000-step trajectories offline and saved the model's state information at each step. For each task, this was repeated across 3,000 different seeds to generate 3,000 different trajectories with 1,000 steps each. 
Note that we only provide the results (Figure \ref{fig:val-losses-Ant} and Table \ref{table:val-losses-Ant}) for \texttt{Ant-v1} here for brevity. The full results and additional analyses are in Section \ref{appendix:robotics} of the appendix.

\begin{figure}[H]
    \centering
    \begin{minipage}{0.48\columnwidth}
        \centering
        \includegraphics[width=\columnwidth]{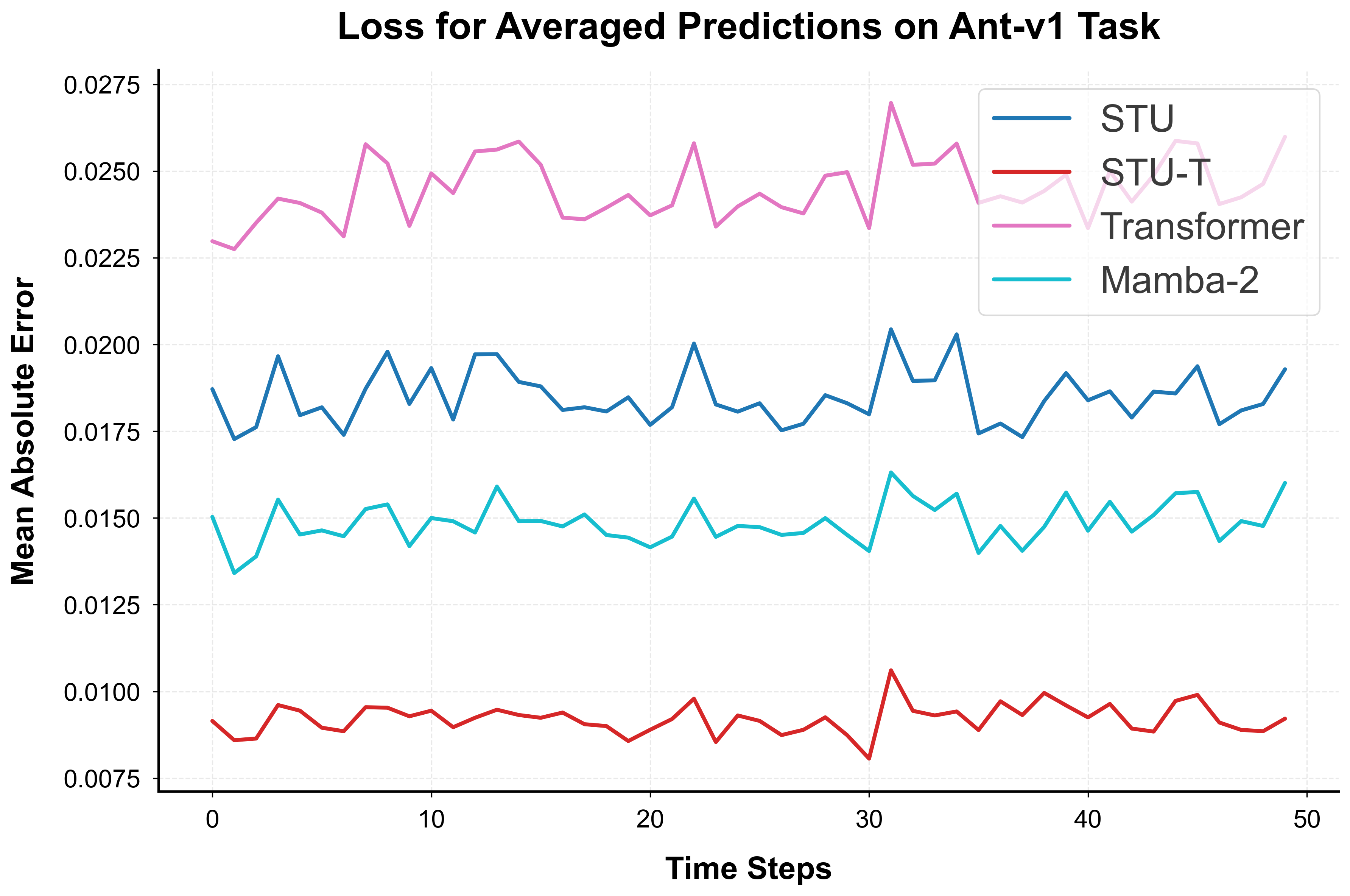}
        \caption{\texttt{Ant-v1} comparative next-step prediction results.}
        \label{fig:pred-losses-Ant}
    \end{minipage}
    \hfill
    \begin{minipage}{0.48\columnwidth}
        \centering
        \includegraphics[width=\columnwidth]{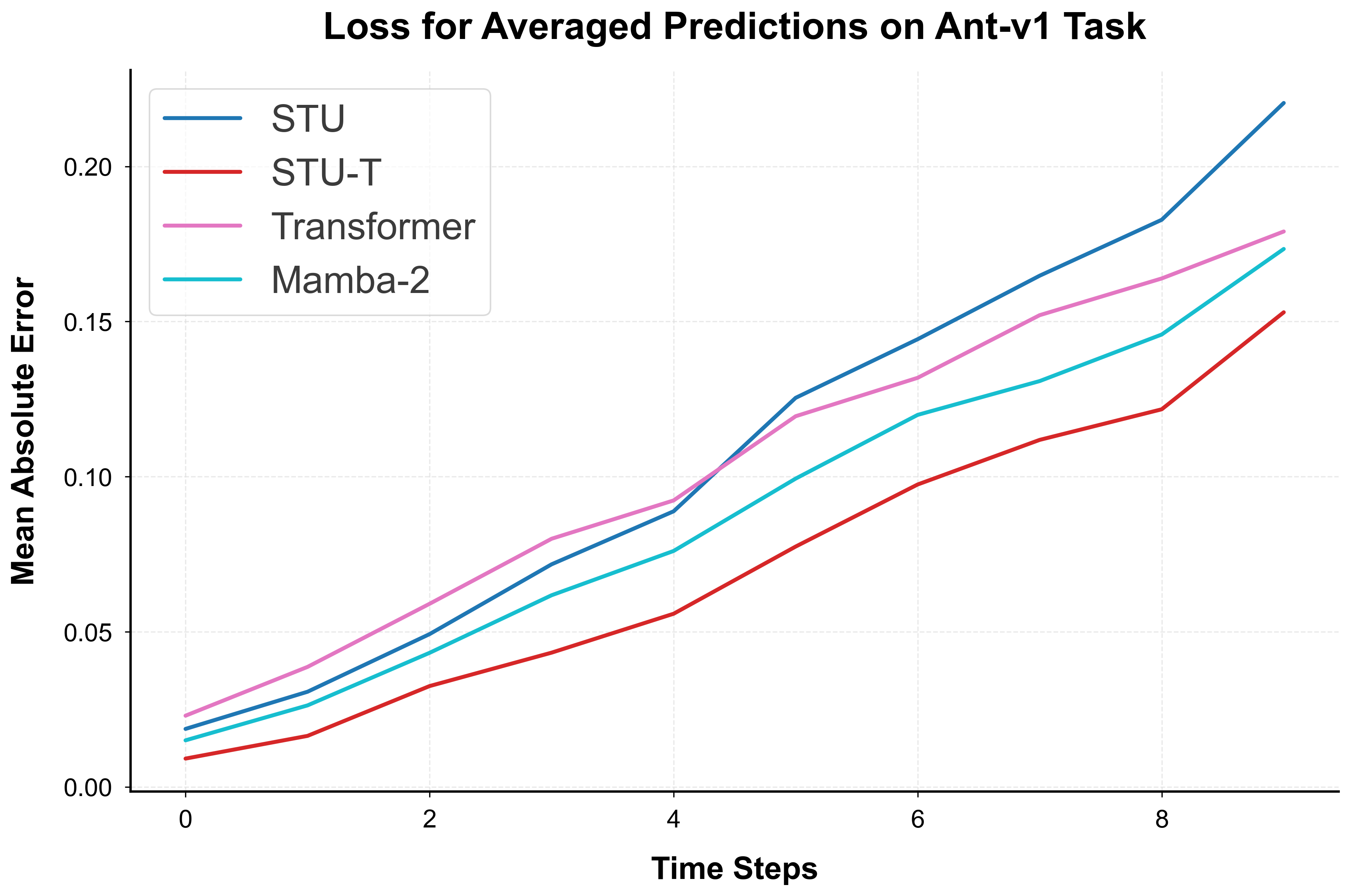}
        \caption{\texttt{Ant-v1} comparative autoregressive next-step prediction results.}
        \label{fig:pred-losses-Ant-ar}
    \end{minipage}
\end{figure}

\textbf{Additional experiments.} To further validate the results, we conduct next-step prediction and autoregressive next-step prediction using the trained models (Figure \ref{fig:pred-losses-Ant} and \ref{fig:pred-losses-Ant-ar}, with losses averaged over 500 predictions for each model). Next-step prediction uses ground truth states and actions from a fixed input window to forecast the next state, mirroring the setting of the evaluation stage during training. Autoregressive next-step prediction, on the other hand, incorporates the model's own past predictions as inputs, allowing errors to accumulate over time.

\begin{table}[H] 
  \centering
  \caption{Model specifications for the \texttt{Ant-v1} Task \label{table:model-specs}.}
  \resizebox{0.99\columnwidth}{!}{%
  \begin{tabular}{@{}lcccccc@{}}
    \toprule
    Model & Parameters & Width/Depth & MLP Scale & Filters & MoE & Time / Train Step \\
    \midrule
    STU & 0.55M & 64/4 & 1 & 16\tablefootnote{Ablation studies~\cite{agarwal2023spectral} on the performance of STU using different numbers of filters ($K$) show that loss stops decaying and plateaus at around $K = 15$.} & False & 11.4ms \\
    STU-T
    % \tablefootnote{One advantage of the tensordot approximation is that it allows us to allocate parameters elsewhere in STU-T while maintaining equal parameter count. For an overview of STU-T's performance with similar model hyperparameters as the other models in this experiment, see Table \ref{table:axes-stu-t}.} 
    & 0.47M & 128/4 & 8 & 16 & False & 11.4ms \\
    Transformer & 0.51M & 64/4 & 6 & - & True & 21.3ms \\
    Mamba-2 & 0.51M & 64/4 & 4 & - & True & 40.5ms  \\
    \bottomrule
  \end{tabular}
  }
\end{table}

\vspace{-1em}

\textbf{Results}. Based on our experiment results, STU-T \textbf{outperforms} all other models. In general, SSMs' subquadratic complexity and ability to theoretically capture infinitely long dependencies~\cite{gu2022efficientlymodelinglongsequences} allow them to excel at these robotics tasks compared to Transformers, with Mamba-2 performing the strongest. More excitingly, the STU-T model outperforms Mamba-2 while being faster, and also shows smoother optimization early in training. The tensordot approximation proves to be particularly effective, reducing parameter count and memory use while maintaining accuracy.

We also observe that STU benefits more from depth, whereas Transformers is affected more by model widths and plateaus quickly beyond two layers. Furthermore, extending STU with sparsely-gated mixture-of-SwiGLUs were less effective, even though SwiGLU configuration~\cite{shazeer2020gluvariantsimprovetransformer} improved performance. We envision there is a ripe field of research in architectural design for the STU, e.g. adaptive online methods and variants of MoE~\cite{hazan2022introduction}, which we leave to future work.

%% file: 03.LLM.tex
\section{Experiments with language modeling}

In this section, we explore the STU's ability for sequence prediction in the context of language modeling. We model our architecture in the style of the LLaMA family of models~\cite{touvron2023llamaopenefficientfoundation}, and we open source a simple, fully distributed large language model (LLM) pretraining pipeline for the community to enjoy and build upon~\cite{flashSTU}.

% Cite ~\cite{flashSTU} after paper review

\subsection{Experiment setup}
\textbf{Data.}
We pretrain on roughly 10B high-quality tokens from FineWeb-Edu~\cite{penedo2024finewebdatasetsdecantingweb}, a large-scale, open source dataset for language modeling. We tokenized the dataset into 95 training shards and 1 validation shard, each containing about 100M tokens, using the \texttt{o200k\_base} tokenizer from the OpenAI ~\texttt{tiktoken}\footnote{\url{https://github.com/openai/tiktoken/}}~ library.

\textbf{General design choices.} 
For each model, we used RMSNorm~\cite{zhang2019rootmeansquarelayer} to pre-normalize the inputs before each attention layer, Flash STU layer, and MLP layer. We followed the standard pre-norm residual connection pattern around each of the sublayers, i.e. the output of each sublayer is \texttt{x + Sublayer(RMSNorm(x))}. To further stabilize training, we capped logits~\cite{bello2017neuralcombinatorialoptimizationreinforcement, grok1, team2024gemma} in each attention layer at $50.0$. Following NanoGPT, we rounded up the vocabulary size of each model to the nearest multiple of 64 in order to use more efficient CUDA kernel pathways. We tied embeddings, and we did not use dropout. 
See Table~\ref{table:model-configs-2B} for the full architectural configurations.

\textbf{Transformer architecture.}
We followed the GPT-2-styled Transformer from NanoGPT~\cite{karpathy2024nanogpt}. We added small optimizations such as FlashAttention-2~\cite{dao2022flashattentionfastmemoryefficientexact}. For the 2B training run, we used the ALiBi~\cite{press2022trainshorttestlong} modification to the attention scores. We used position interpolation~\cite{chen2023extendingcontextwindowlarge} to help the model ``length generalize" beyond its context window size from training. It has been shown that position interpolation works even with ALiBi~\cite{alkhateeb2023positioninterpolationimprovesalibi}, so we used an interpolation factor of $0.25$, extending the model's effective context window size at inference time by a factor of $4$. For the 500M runs, we used RoPE~\cite{su2024roformer}. In terms of performance, we find that there is no significant performance difference between the two positional embedding schemes. We used highly optimized Triton~\footnote{\url{https://openai.com/index/triton/}} kernels~\cite{hsu2024ligerkernel} for the implementation of our MLP (SwiGLU), RMSNorm, and loss function.

\textbf{Flash STU architecture.}
We augmented the STU-T model with Flash FFT~\cite{fu2023flashfftconvefficientconvolutionslong} for efficient convolutions. We found that the tensordot approximation was necessary to scale up STU-based models, as it was difficult to scale beyond 1B parameters without experiencing frequent out-of-memory (OOM) errors. Following the work from the state space literature~\cite{dao2024transformersssmsgeneralizedmodels, waleffe2406empirical} relating to the role of attention layers, we opted to use a simple \textit{hybrid} model architecture with alternating layers consisting of STU-T and local attention, both followed by an MLP layer. To maintain efficiency, we used the same optimized Triton kernels from the Transformer, and we replaced global attention with sliding window attention~\cite{beltagy2020longformerlongdocumenttransformer} with window size being one-eighth of the sequence length.

\begin{figure}[H]
\centering
\scalebox{0.5}{
\begin{tikzpicture}[
    block/.style={
        rectangle, 
        draw=black!50, 
        fill=blockColor!20, 
        text width=2.2cm, 
        text centered, 
        rounded corners=0.2cm, 
        minimum height=1.8em,
        font=\sffamily\small\bfseries,
        drop shadow={shadow xshift=0.3mm, shadow yshift=-0.3mm, opacity=0.2}
    },
    smallblock/.style={
        rectangle, 
        draw=black!50, 
        fill=blockColor!20, 
        text width=1.8cm, 
        text centered, 
        rounded corners=0.2cm, 
        minimum height=1.5em,
        font=\sffamily\footnotesize\bfseries,
        drop shadow={shadow xshift=0.2mm, shadow yshift=-0.2mm, opacity=0.2}
    },
    circ/.style={
        circle, 
        draw=black!50, 
        fill=white,
        minimum size=0.6cm,
        font=\sffamily\footnotesize\bfseries,
        drop shadow={shadow xshift=0.2mm, shadow yshift=-0.2mm, opacity=0.2}
    },
    line/.style={
        draw=black!80,
        -stealth,
        line width=0.8pt,
        shorten >=2pt,
        shorten <=2pt
    },
    skip line/.style={
        draw=red!80,
        -stealth,
        dashed,
        line width=0.8pt,
        shorten >=3pt,
        shorten <=3pt
    },
]
    % Color definitions (refined)
    \definecolor{inputColor}{RGB}{70,130,180}
    \definecolor{stuColor}{RGB}{60,179,113}
    \definecolor{outputColor}{RGB}{255,140,0}
    \definecolor{borderColor}{RGB}{47,79,79}
    \definecolor{annotationColor}{RGB}{100,100,100}

    % Calculate center point
    \pgfmathsetmacro{\centerX}{0}

    % Input
    \node[block, fill=inputColor!30] (input) at (\centerX,0) {Input};

    % STU portion
    \node[smallblock] (rn1) [below=0.8cm of input] {RMSNorm};
    \node[block, fill=stuColor!30] (stu) [below=0.6cm of rn1] {STU-T / SWA$^{\dagger}$};
    \node[circ] (add4) [below=0.6cm of stu] {+};
    \node[smallblock] (rn2) [below=0.6cm of add4] {RMSNorm};
    \node[block] (mlp) [below=0.6cm of rn2] {MLP};
    \node[circ] (add1) [below=0.6cm of mlp] {+};

    % Output
    \node[block, fill=outputColor!30] (output) [below=0.8cm of add1] {Output};
    
    % Connections
    \path [line] (input) -- (rn1);
    \path [line] (rn1) -- (stu);
    \path [line] (stu) -- (add4);
    \path [line] (add4) -- (rn2);
    \path [line] (rn2) -- (mlp);
    \path [line] (mlp) -- (add1);
    \path [line] (add1) -- (output);
    
    % Skip Connections
    \path [skip line] (input) -| ($(stu.west)+(-0.6,0)$) |- ($(add4.west)+(0,0.1)$);
    \path [skip line] (add4) -| ($(mlp.west)+(-0.6,0)$) |- ($(add1.west)+(0,0.1)$);

    % Block outlines
    \node[fit=(rn1) (add1), inner sep=0.3cm, draw=stuColor, thick, rounded corners=0.3cm] (stu_block) {};

    % Nx annotation
    \node[right=0.05cm of stu_block, font=\sffamily\small\bfseries, black] (nx) {$\times N$};
    
    % Add balancing space on the left
    \node[left=0.6cm of stu_block] (balance) {};
    
    % Add strong border around the entire architecture
    \node[fit=(input) (output) (stu_block) (nx) (balance), inner sep=0.5cm, draw=borderColor, line width=1.2pt, rounded corners=0.4cm] (border) {};

    % Add title
    % \node[above=0.4cm of border.north, font=\sffamily\normalsize\bfseries] {Flash STU Model Architecture};

\end{tikzpicture}
}
\caption{Flash STU model architecture, alternating between STU-T and (sliding window) attention$^{\dagger}$.}
\label{fig:flash-stu-model-architecture}
\end{figure}

\begin{table*}[t]
\centering
\caption{Performance comparison of Flash STU and Transformers across various language modeling evaluations, such as MMLU~\cite{hendrycks2021measuringmassivemultitasklanguage}, HellaSwag~\cite{zellers2019hellaswagmachinereallyfinish}, PIQA~\cite{bisk2019piqareasoningphysicalcommonsense}, BoolQ~\cite{clark2019boolqexploringsurprisingdifficulty}, WinoGrande~\cite{sakaguchi2019winograndeadversarialwinogradschema}, CommonsenseQA~\cite{talmor2019commonsenseqaquestionansweringchallenge}, OpenBookQA~\cite{mihaylov2018suitarmorconductelectricity}, and ARC~\cite{clark2018thinksolvedquestionanswering}.\label{table:model-performance}}
\label{table:model-performance-comparison}
\resizebox{\textwidth}{!}{%
\begin{tabular}{@{}lccccccccccc@{}}
\toprule
\textbf{Model} & \textbf{MMLU} & \textbf{Hella.} & \textbf{PIQA} & \textbf{BoolQ} & \textbf{Wino.} & \textbf{CSQA} & \textbf{OBQA} & \textbf{ARC-e} & \textbf{ARC-c} & \textbf{Average} \\ 

 & (acc ↑) & (acc\_n ↑) & (acc\_n ↑) & (acc ↑) & (acc ↑) & (acc ↑) & (acc\_n ↑) & (acc ↑) & (acc\_n ↑) & (↑) \\ 
\midrule
Flash STU 550M& \textbf{26.31} & \textbf{28.64} & \textbf{60.94} & \textbf{52.23} & \textbf{50.67} & 19.66 & 26.00 & \textbf{45.16} & \textbf{23.63} & \textbf{37.03} \\ 
Mamba-2 Hybrid 546M& 25.82 & 28.51 & 57.62 & 51.87 & 49.09 & 18.67 & \textbf{27.60} & 44.36 & 22.53 & 36.23 \\ 
Mamba-2 561M& 24.15 & 26.83 & 57.62 & 46.54 & 50.12 & \textbf{20.23} & 26.60 & 44.95 & 23.38 & 35.60 \\ 
Transformer 564M& 25.16 & 26.85 & 56.53 & 51.41 & 50.51 & 19.82 & 25.00 & 38.64 & 21.25 & 35.02 \\ 
\bottomrule
\end{tabular}
}
\end{table*}

\vspace{-1em}
\textbf{Training.}
We used fused AdamW~\cite{loshchilov2019decoupledweightdecayregularization} with default PyTorch hyperparameters, and we set the maximum learning rate and minimum learning rate to $3.0 \times 10^{-4}$ and $3.0 \times 10^{-5}$, respectively, following~\cite{hoffmann2022training, rae2021scaling}. We used a linear decay with warmup learning rate schedule~\cite{defazio2023whenmuchadaptivelearning} and allocated $10\%$ of training steps to warm up. Each model was trained for 16 hours across $8\times$H100 80GB HBM3 GPUs using the Fully Sharded Data Parallel (FSDP) framework from PyTorch~\cite{zhao2023pytorch}.

\subsection{Results}

\begin{figure}[H]
    \centering
    \includegraphics[width=0.85\columnwidth]{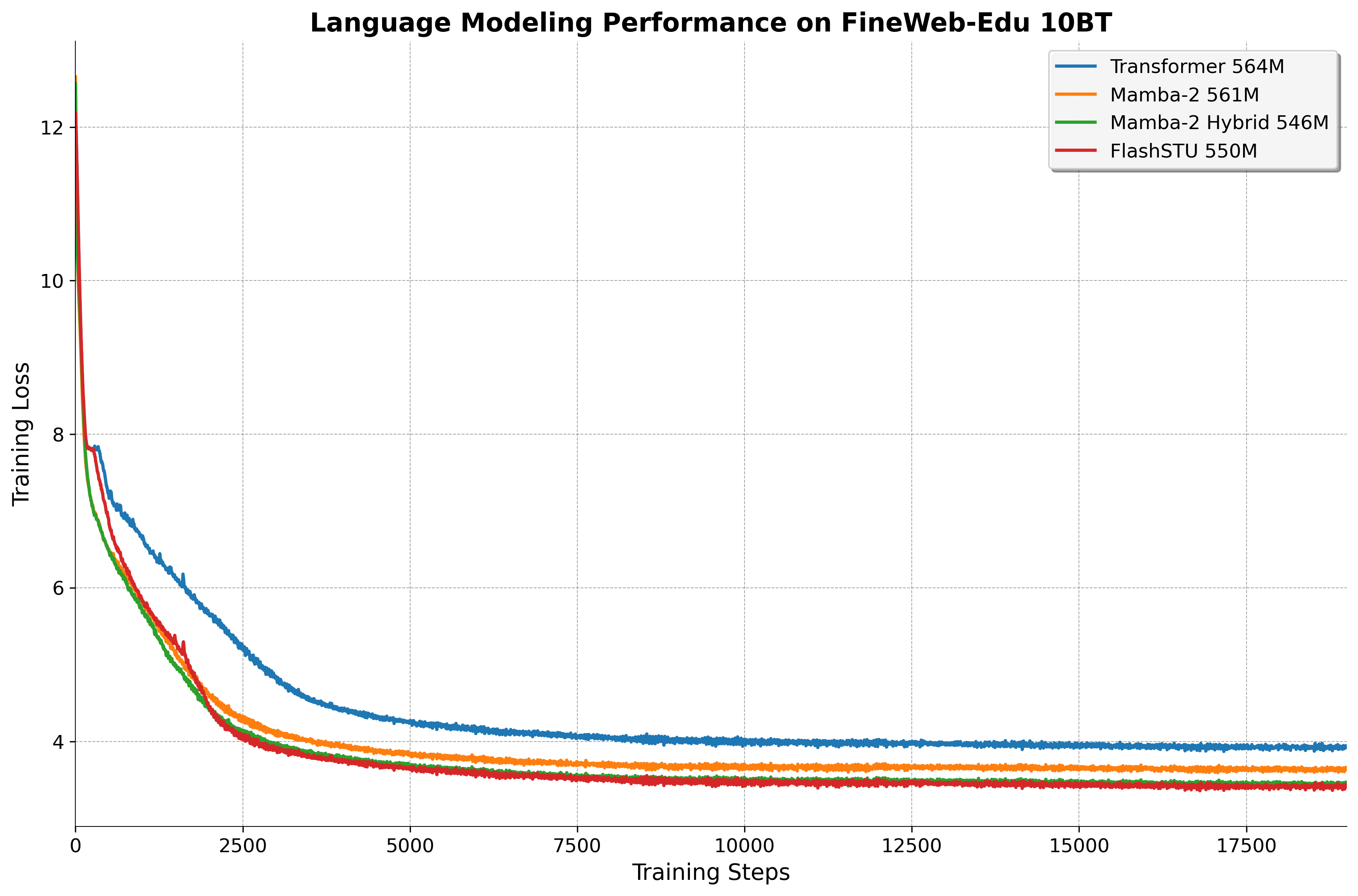}
    \caption{Training curves for Flash STU and baseline models at different scales.}
    \label{fig:all-models}
\end{figure}

\begin{figure}[H]
    \centering
    \includegraphics[width=0.85\columnwidth]{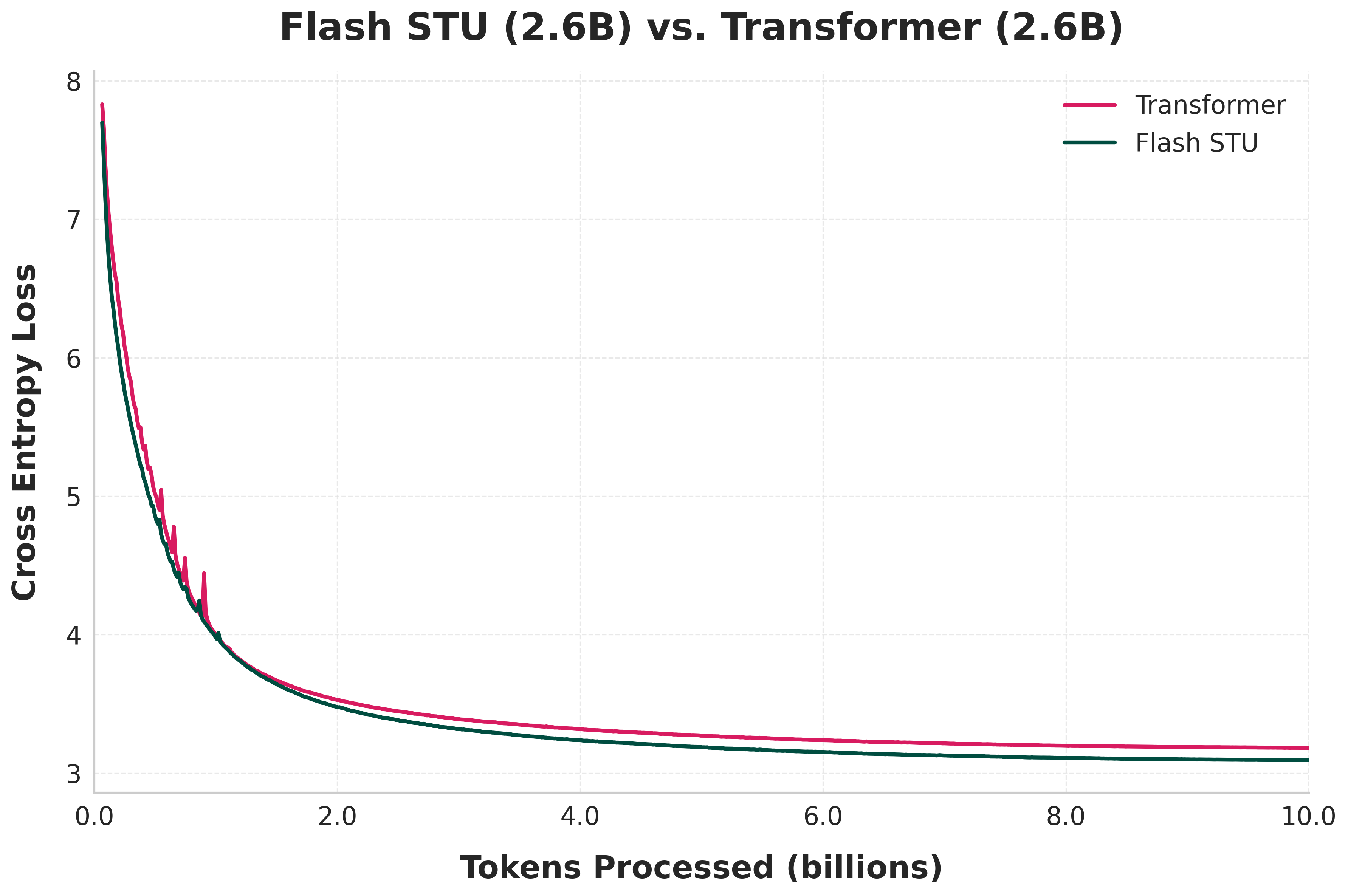}
    \caption{Comparison of Flash STU and Transformer models on validation set.}
    \label{fig:comparison}
\end{figure}

First, we investigate whether Flash STU can scale to billion-parameter models without memory or stability issues. This crucial scaling property~\cite{kaplan2020scaling} is the determining factor of model architecture adoption. After verifying successful training at the 2B scale, we focus our benchmark evaluations on compute-optimal model sizes for 10B training tokens, which is approximately 500M parameters~\cite{hoffmann2022training}. 
See Table \ref{table:model-configs-500M} for model configurations.
Our main result is that the Flash STU architecture \textbf{outperforms} the Transformer, Mamba-2, and even a Mamba-2 hybrid\footnote{For simplicity, we alternate the Mamba and attention layers as commonly seen in the literature~\cite{lieber2024jamba, ren2024samba}. See~\cite{waleffe2406empirical} for a more fine-grained approach to the hybridization of the Mamba architecture.}. Our results are presented in Figures \ref{fig:all-models}, \ref{fig:comparison} and Tables \ref{table:model-performance}, \ref{table:500m-performance}.

\begin{table}[ht]
    \centering
    \caption{Performance of 500M models.\label{table:500m-performance}}
    \resizebox{0.55\columnwidth}{!}{%
    \begin{tabular}{@{}lc@{}}
    \toprule
    Model & Validation Loss \\
    \midrule
    Flash STU & \textbf{3.40} \\
    Mamba-2 Hybrid & 3.44 \\
    Mamba-2 & 3.63 \\
    Transformer & 3.92 \\
    \bottomrule
    \end{tabular}
    }
\end{table}

\begin{table}[ht]
    \centering
    \caption{Performance of 2B models.}
    \resizebox{0.95\columnwidth}{!}{%
    \begin{tabular}{@{}lcccc@{}}
    \toprule
    Model & Validation Loss & Perplexity & Mean Time/Step \\
    \midrule
    Flash STU & \textbf{3.09} & \textbf{22.08} & \multirow{2}{*}{$\approx$ 3.0s} \\
    Transformer & 3.18 & 24.11 & \\
    \bottomrule
    \end{tabular}
    }
\end{table}

Qualitatively, we found that the Transformer was more sensitive to hyperparameter tuning, e.g. learning rate. Additionally, the Transformer was more prone to spikes in its loss compared to Flash STU despite our best attempts to stabilize training.

We also note that although our Flash STU implementation is fairly optimized, it is not yet at the level at which the community has optimized the Transformer model in recent years. Similarly, while Flash STU has the edge in terms of asymptotic time complexity, the Transformer is heavily optimized for matrix multiplication units on modern accelerator hardware, which generally translates to superior wall-clock performance in practice. However, we notably found that the average time per step was approximately around the same for both models. Thus, we believe that the Flash STU model demonstrates great promise for future improvements given that it outperformed the Transformer, while having an asymptotically better time complexity, as well as Mamba and its attention hybrid variant.

%% file: 04_future_work.tex
\section{Conclusion and future directions}
This paper presented a methodological extension and experimental evaluation of the theoretically-founded Spectral Transformers. We evaluate our hybrid Flash STU architecture on three modalities: synthetic data, robotics control, and language modeling. We notably found that the STU is a viable and competitive alternative to other popular sequence models such as the Transformer and various state space models. %at \cite{flashSTU}. 

%% file: 06.acks.tex
\section{Acknowledgements}
We thank Naman Agarwal, Xinyi Chen, and Daniel Suo for helpful conversations. 
The authors were funded in part by Elad Hazan's grants from the Office of Naval Research and the National Science Foundation. We would like to give special thanks to Princeton Research Computing\footnote{\url{https://researchcomputing.princeton.edu/}} and Princeton Language and Intelligence\footnote{\url{https://pli.princeton.edu/}} (PLI) for supplying us with the compute that makes our experiments possible.

%% file: appendix.tex
\appendix
\subsection{Spectral filters}
\label{appendix:spectral-filters}
Following~\cite{agarwal2023spectral}, we construct spectral filters by extracting the top $K$ eigenvectors from the Hankel matrix:
\begin{equation}
Z(i, j) = \frac{2}{(i + j)^3 - (i + j)},
\end{equation}
\noindent
These top $K$ eigenvectors are then scaled by their corresponding eigenvalues, each raised to the $\frac{1}{4}$-th power. These scaled eigenvectors serve as our spectral filters and are used to convolve inputs into the spectral basis.

\vspace{0.5cm}
\begin{figure}[H]
    \centering
    \includegraphics[width=0.65\textwidth]{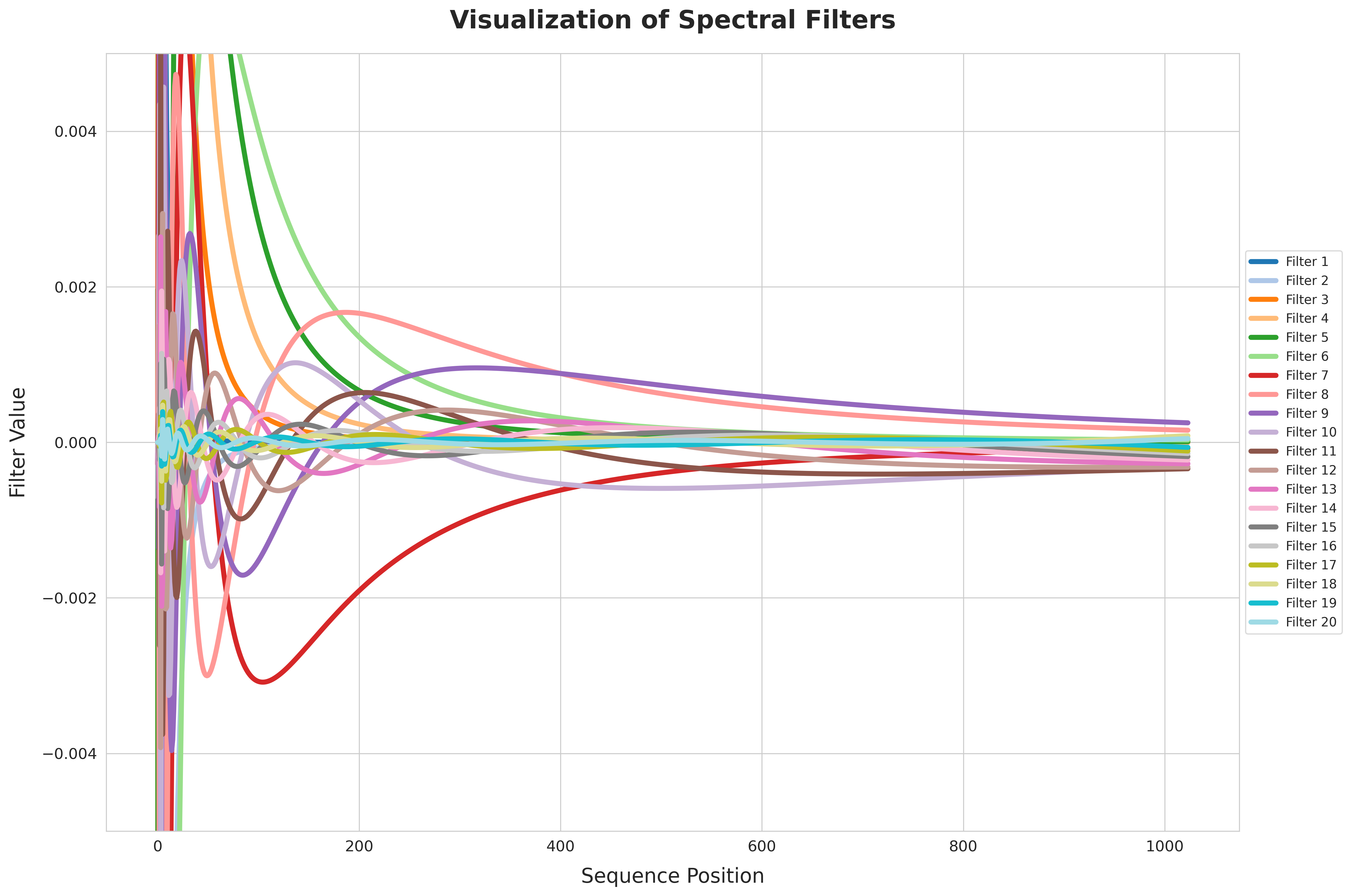}
    \caption{
        Visualization of the first $20$ spectral filters used in our STU-based models.
    }
\end{figure}

%%%%%%%%%%%%%%%%%%%%%%%%%%%%%%%%%%%%%%%%%%%%%%%%%%%%%%%%%%%%%%%%%%%%%%%%%%

\subsection{Additional synthetic experiments}
\label{appendix:synthetics}
\textbf{Extra experimental details.} The synthetic experiments were all run on a single A100 GPU, with each trial for each model taking around a minute to complete. The STU implementation followed Figure \ref{fig:arch} exactly, with the tensordot approximation replacing the STU layer in the STU-T implementation. The softmax attention baseline is FlashAttention-2~\cite{dao2022flashattentionfastmemoryefficientexact}, and for S4 we used the official implementation\footnote{\url{https://github.com/state-spaces/s4/blob/main/models/s4/s4d.py/}}~\cite{gu2022efficientlymodelinglongsequences}. Whenever we introduce nonlinearities in the synthetic experiments, it is always ReLU; we leave a close investigation of multiplicative gating nonlinearities like GLU to future work. We used a learning rate of 0.0024 for Adam, which was tuned to be best for the attention and Mamba-2 baselines (as they were the least robust to learning rate). Anecdotally, we note that STU seems stable with much larger learning rates, which we hope practitioners will be able to take advantage of.

\begin{figure}[H]
\centering
    \begin{minipage}{0.24\textwidth}
        \centering  
        \includegraphics[width=\textwidth]{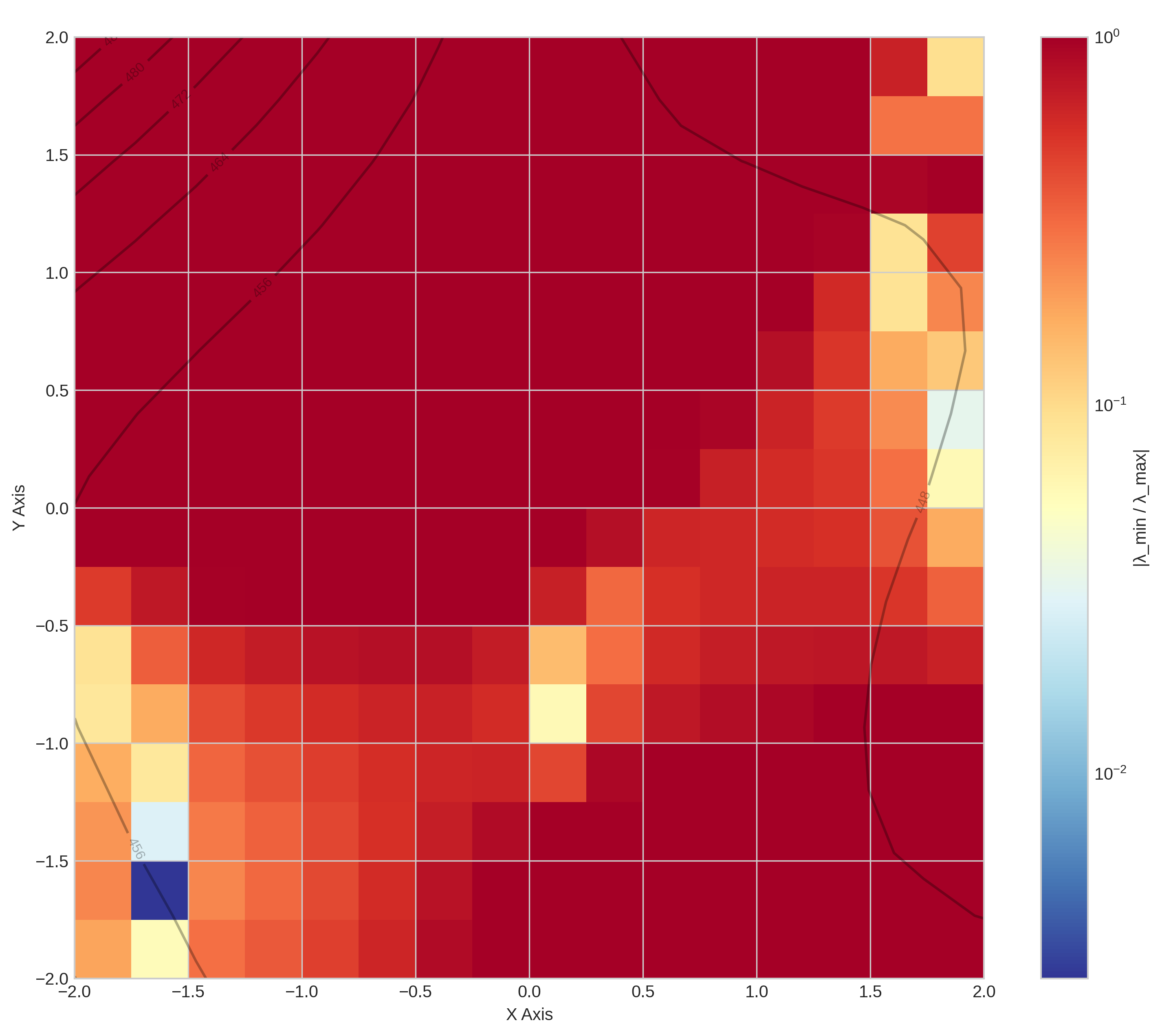}
        (a) STU layer.
    \end{minipage}
    \hfill
    \begin{minipage}{0.245\textwidth}
        \centering
        \includegraphics[width=\textwidth]{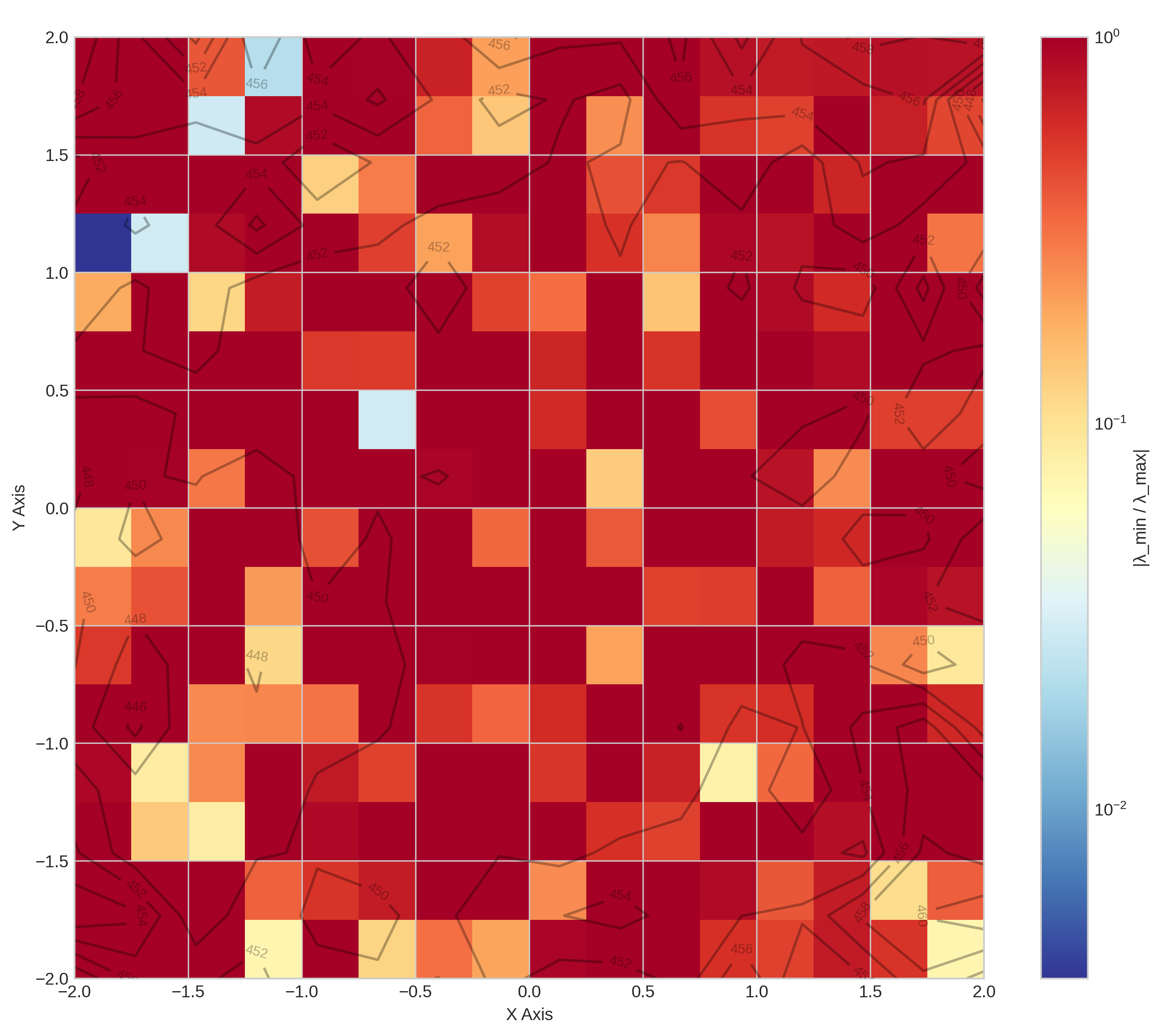}
        (b) S4 layer.
    \end{minipage}
    \hfill
    \begin{minipage}{0.24\textwidth}
        \centering  
        \includegraphics[width=\textwidth]{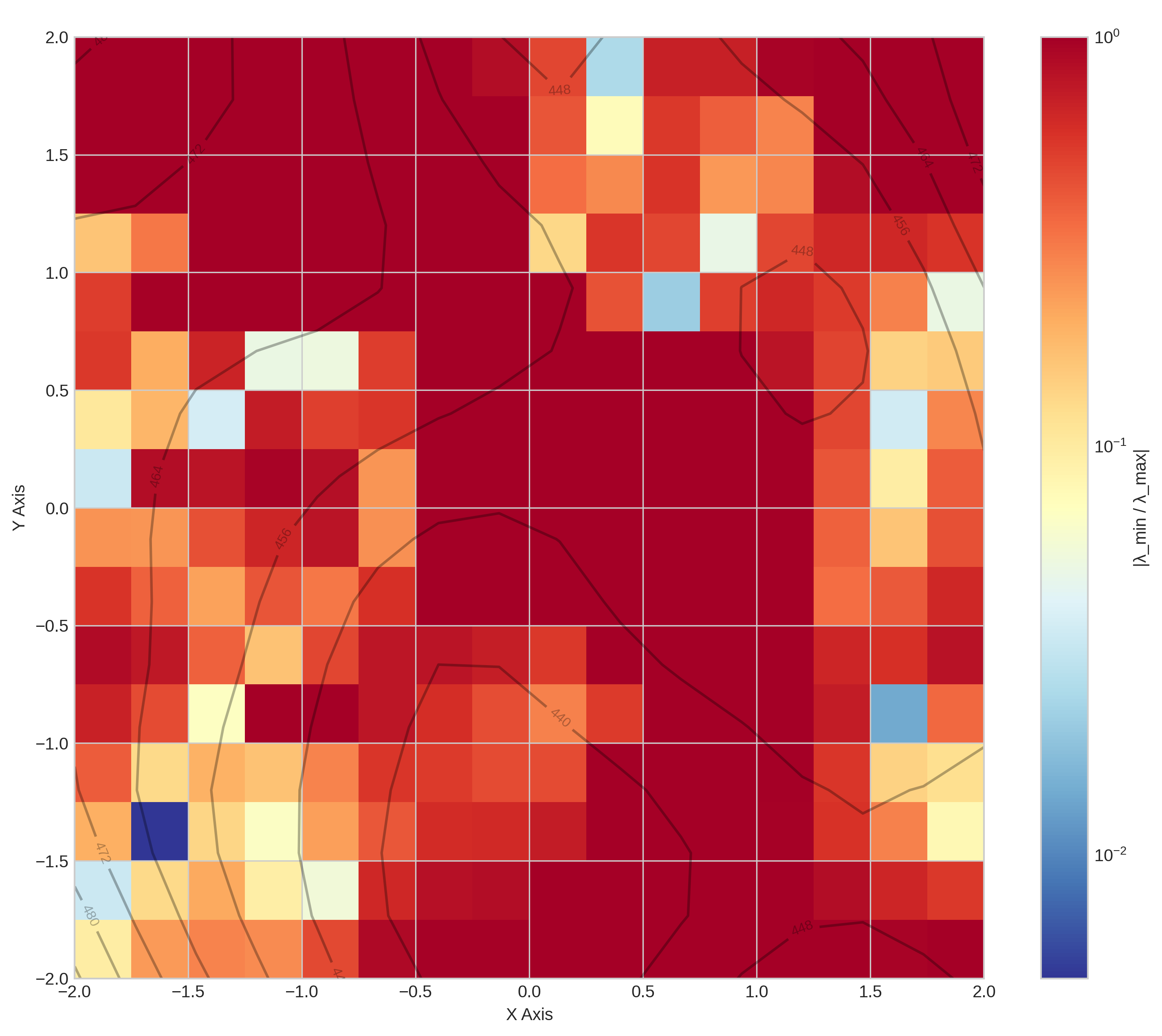}
        (c) Mamba-2 layer.
    \end{minipage}
    \hfill 
    \begin{minipage}{0.24\textwidth}
        \centering
        \includegraphics[width=\textwidth]{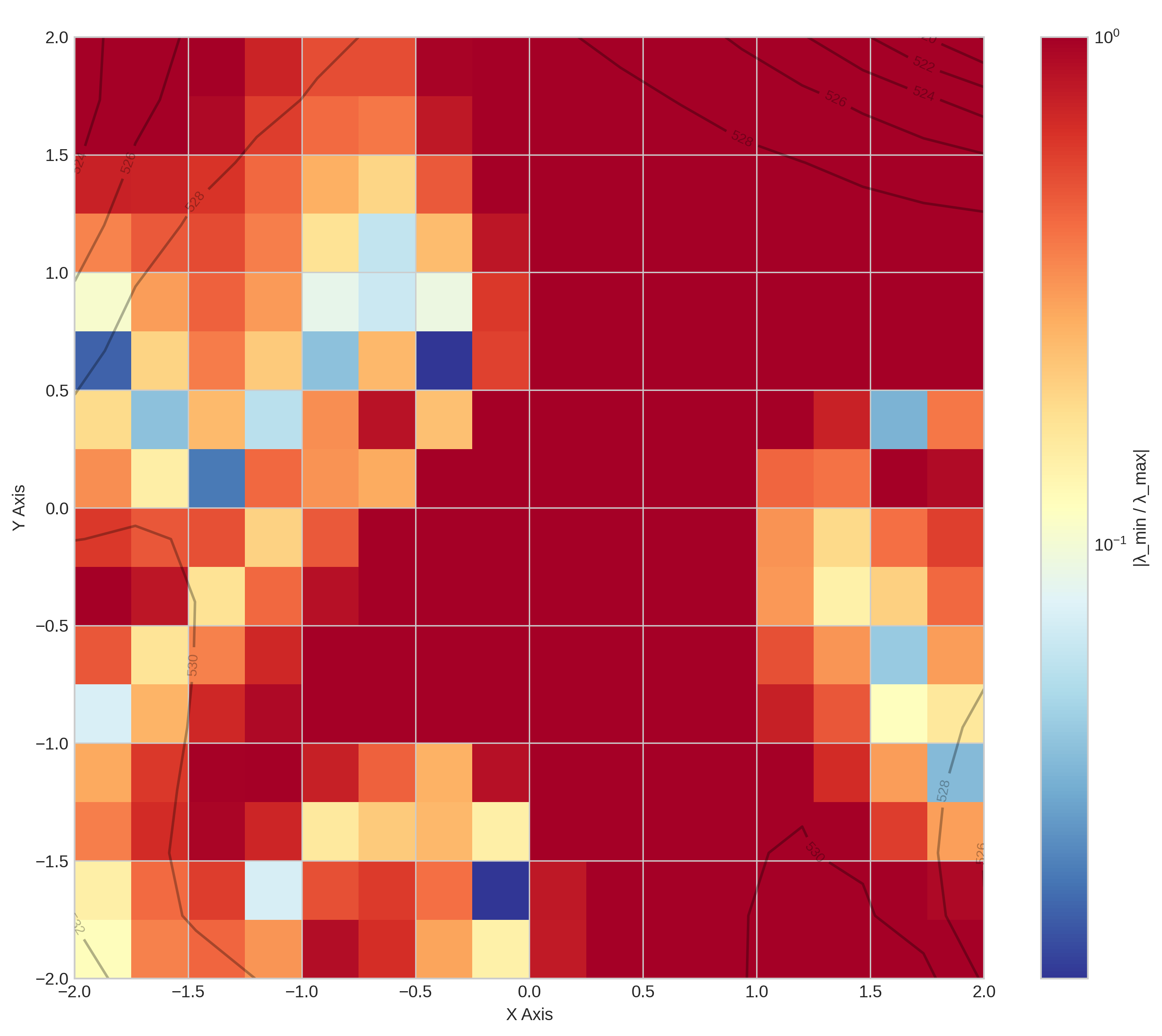}
        (d) Attention layer.
    \end{minipage}

\caption{Heat maps of the ratio $\left|\lambda_{\min} / \lambda_{\max}\right|$ of (an estimate of a dimensionality-reduced version of) the loss Hessian.}
\label{fig:synthetics-eigvals-heatmaps}
\end{figure}

\textbf{Local curvature of the loss.} 
In Figure \ref{fig:synthetics-eigvals-heatmaps}, eigenvalue ratios $|\lambda_{\min} / \lambda_{\max}|$ of the loss Hessian are plotted after 10 steps of training on the LDS task from the main paper. A ratio near $1.0$ means the curvature is similar in all directions, while a ratio near $0.0$ means the curvature is stronger in one direction. 
% We use absolute value because we are interested in the magnitude of the curvature. 
The contour lines overlaid on the heat map show the loss landscape, allowing us to visualize how this local geometry relates to the overall loss surface. Red areas represent more spherical curvature, and blue areas represent more elongated curvature. Movement along the $x$ and $y$ axes corresponds to local movement in parameter space.

For STU, we see very smooth behavior and obvious directions of progress toward well-conditioned areas, whereas the loss landscape is more complicated for Mamba-2, attention, and especially S4.
\vspace{0.5cm}

\textbf{Selective copy.}
To better understand how the STU layer performs on different types of small sequence prediction problems, we also experiment on the selective copy task~\cite{arjovskyshah}.
% and prediction of the mode of the tokens in the context.  
The selective copy task requires the model to recall (in order) a sequence of a fixed number of tokens, which are sampled uniformly from a vocabulary and distributed randomly throughout the context of otherwise \verb|blank| tokens.
% By contrast, mode prediction simply requires the model to output the mode of a sequence of tokens sampled uniformly from a vocabulary\footnote{As far as we can tell, discrete counting problems are not frequently used to study the behaviors of sequence models in any existing literature. We think it is an elegant way to probe a model's ability to nonlinearly synthesize very simple single-token information, as it is clear how the difficulty of the problem scales with sequence length and vocabulary size. It may also lend itself to strong theoretical analyses on account of its simplicity, and there is clear structure with which to study generalization to new tokens or seqence lengths.}. 
% \\
% \\ 
Doing this successfully requires the ability to synthesize single-token resolution information across the whole sequence. 
% The accuracy during training is plotted in Figure \ref{fig:synthetics-selectivecopy}.
% Speaking broadly, an LDS requires a linear aggregation of the context, while induction heads requires only local understanding of the context. Selective copy combines these principles, testing a model's ability to synthesize single-token-precision information across the whole sequence. 
% Training results are presented in Figures \ref{fig:synthetics-mode} and \ref{fig:synthetics-selectivecopy}, respectively.
% To complement the experiments in the main paper, we run additional synthetic tasks with slightly different properties. 
% The experiments in the main paper are almost a minimal requirement for a sequence prediction layer in modern deep learning

\begin{figure}[H]
    % \begin{minipage}{0.45\textwidth}
    %     \centering  \includegraphics[width=\textwidth]{figures/mode_accs.png}
    %     \caption{Prediction accuracy for the mode of the context during training.
    % }
    %     \label{fig:synthetics-mode}
    % \end{minipage}
    % \hfill
    % \begin{minipage}{0.45\textwidth}
        \centering
\includegraphics[width=0.55\textwidth]{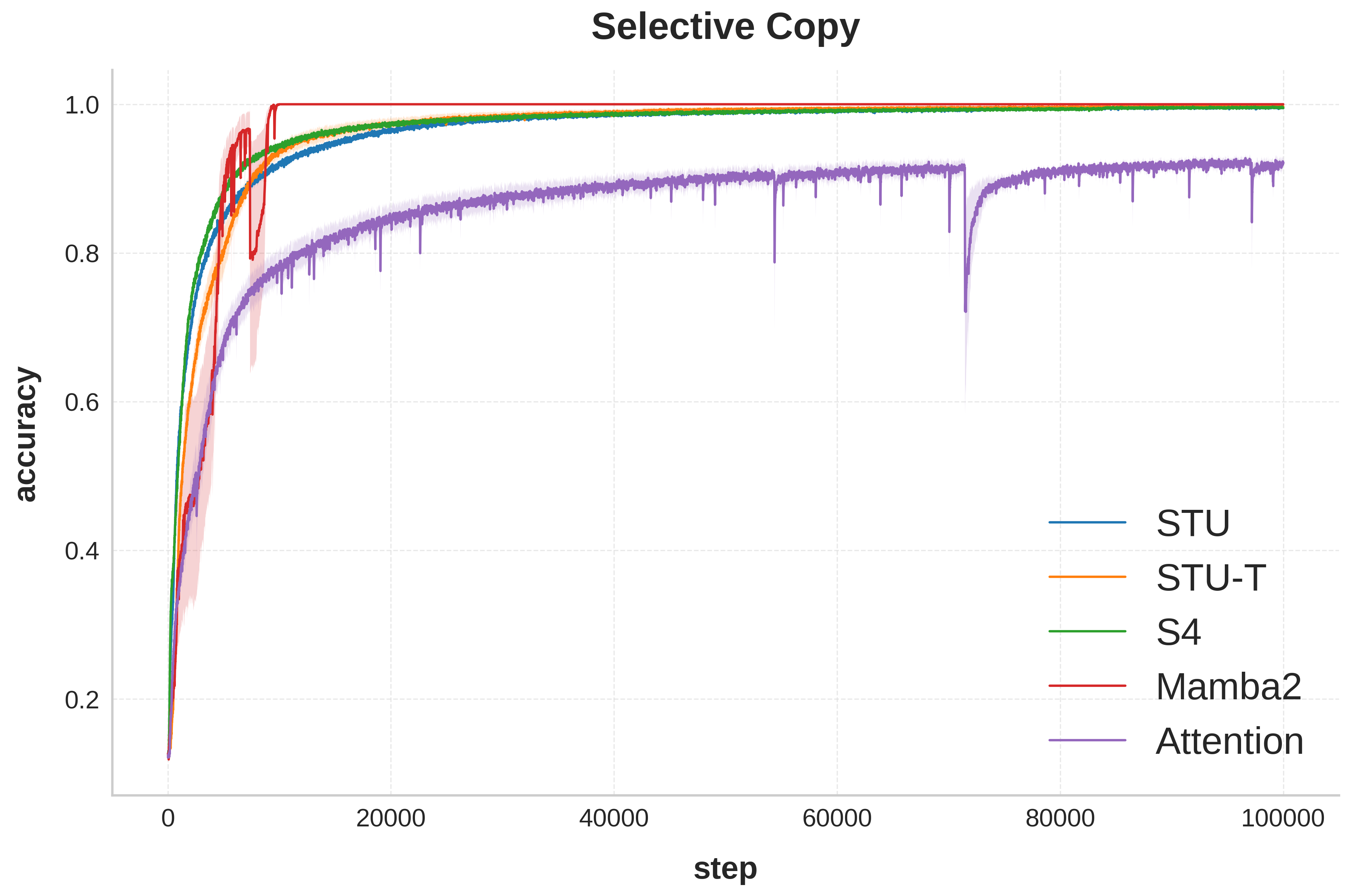}
\caption{
    Prediction accuracy for repeating the non-blank tokens in the context during training.
% \evan{context len = 32, batch size = 128, num steps = 10000, vocab size = 5, depth=2 with mlp, opt = standard adam 1e-3}
}
    \label{fig:synthetics-selectivecopy}
    % \end{minipage}
\end{figure}

\textbf{Experiment details and conclusions.}
As before, we train two-layer models with MLP layers and nonlinearities in between. The models have width 32 and are trained with a tuned Adam optimizer. 
% Both tasks are constructed with a sequence length of 128 from a small vocabulary of 10 tokens. These tasks, 
The selective copy dataset is constructed with a sequence length of 128 tokens with a vocabulary of 10 tokens to copy each training example. This task, like those in the main paper, is in the regime where models are wide enough to memorize the vocabulary but not the whole context. We leave the study of phenomena under different width and depth scalings to future work.

As we see in Figure \ref{fig:synthetics-selectivecopy}, Mamba-2 appears to converge to a solution with qualitatively different behavior than the other SSM methods. Put loosely, we suspect that Mamba finds a nonlinear, selective solution while STU and S4 find a linear one which also solves this instance of the task. It is up to architecture designers to choose whether to value a more expressive layer (like Mamba-2) or a simpler, more parameter-efficient, and easier to optimize layer (like STU-T) for use in large models. This experiment serves as a good reminder that this tradeoff looks different depending on the task and scale. Lastly, similarly to the associative recall task, we find that our attention model seems to have difficulty optimizing even though we tune the learning specifically for attention. 
% This experiment serves as a good reminder that some synthetic tasks are more informative than others, and that they are not make-or-break predictors for model performance on other problems or at scale.
% \\ On mode prediction, we see the same surprising results as on induction heads and associative recall: SSM models perform learn the quickest, with STU leading the way. However, the S4 model dominates on selective copy, while STU-T appears able to match the attention baseline. All models eventually converge to perfect accuracy and are always able to express a solution for this task, but this time it is S4 that appears to have the easiest optimization. % \evan{does this result make sense? its the only place STU doesnt win} 

To see empirical results of STU and S4 on Long Range Arena benchmark, see Table 1 of~\cite{agarwal2023spectral}.

%%%%%%%%%%%%%%%%%%%%%%%%%%%%%%%%%%%%%%%%%%%%%%%%%%%%%%%%%%%%%%%%%%%%%%%%%%

\subsection{Additional model architectures}
\label{appendix:architectures}
Figure~\ref{fig:model-architectures} shows the model architectures used for the robotics experiments
in Section \ref{robotics-experiments}. We ran small ablation studies and found that global skip connections often improved performance in our MuJoCo robotics experiment setting. In non-hybrid STU architectures, we found that performance decreases when using an MoE architecture, so we used a simple gated MLP instead.

% Color definitions (refined)
\definecolor{inputColor}{RGB}{70,130,180}
\definecolor{stuColor}{RGB}{60,179,113}
\definecolor{outputColor}{RGB}{255,140,0}
\definecolor{borderColor}{RGB}{47,79,79}
\definecolor{annotationColor}{RGB}{100,100,100}

\begin{figure}[H]
\centering
\begin{minipage}[t]{0.3\textwidth}
\centering
\begin{tikzpicture}[scale=0.5,
    block/.style={
        rectangle, 
        draw=black!50, 
        fill=blockColor!30, 
        text width=1.8cm, 
        text centered, 
        rounded corners=0.2cm, 
        minimum height=1.8em,
        font=\sffamily\scriptsize\bfseries,
        drop shadow={shadow xshift=0.3mm, shadow yshift=-0.3mm, opacity=0.3}
    },
    smallblock/.style={
        rectangle, 
        draw=black!50, 
        fill=smallBlockColor!30, 
        text width=1.8cm, 
        text centered, 
        rounded corners=0.15cm, 
        minimum height=1.4em,
        font=\sffamily\scriptsize\bfseries,
        drop shadow={shadow xshift=0.2mm, shadow yshift=-0.2mm, opacity=0.3}
    },
    circ/.style={
        circle, 
        draw=black!50, 
        fill=white,
        minimum size=0.6cm,
        font=\sffamily\scriptsize\bfseries,
        drop shadow={shadow xshift=0.1mm, shadow yshift=-0.1mm, opacity=0.3}
    },
    line/.style={
        draw=black!70,
        -latex',
        line width=0.8pt,
        shorten >=1pt,
        shorten <=1pt
    },
    skip line/.style={
        draw=red!70,
        -latex',
        dashed,
        line width=0.8pt,
        shorten >=2pt,
        shorten <=2pt
    },
]
    % STU-based Architecture (content unchanged, but with adjusted spacing)
    \node[block, fill=inputColor!30] (input) at (0,0) {Input};
    \node[block] (rn1) [below=0.7cm of input] {RMSNorm};
    \node[block] (stu) [below=0.4cm of rn1] {STU};
    \node[circ] (add2) [below=0.4cm of stu] {+};
    \node[block] (rn2) [below=0.4cm of add2] {RMSNorm};
    \node[block] (mlp) [below=0.4cm of rn2] {MLP};
    \node[circ] (add1) [below=0.4cm of mlp] {+};
    \node[fit=(rn1) (add1), inner sep=0.2cm, draw=transformerColor, thick, rounded corners=0.3cm] (block) {};
    
    \node[block, fill=outputColor!30] (output) [below=0.7cm of block] {Output};
    
    % Connections
    \path [line] (input) -- (rn1);
    \path [line] (rn1) -- (stu);
    \path [line] (stu) -- (add2);
    \path [line] (add2) -- (rn2);
    \path [line] (rn2) -- (mlp);
    \path [line] (mlp) -- (add1);
    \path [line] (add1) -- (output);
    % Skip Connections with vertical separation
    \path [skip line] (input) -| ($(block.east)+(0,1.5)$) |- ($(add2.east)$);
    \path [skip line] ($(add2.west)$) -| ($(block.west)-(0,0.65)$) |- (add1);
    
    \node[right=0.05cm of block, font=\sffamily\scriptsize\bfseries, transformerColor] (nx) {$\times N$};
    
    \node[fit=(input) (output) (block) (nx), inner sep=0.3cm, draw=borderColor, line width=1pt, rounded corners=0.5cm] (border) {};
\end{tikzpicture}
\vspace{0.2em}\\
(a) STU-based architecture.
\end{minipage}
\hfill
\begin{minipage}[t]{0.38\textwidth}
\centering
\begin{tikzpicture}[scale=0.5,
    % Define styles
    block/.style={
        rectangle, 
        draw=black!50, 
        fill=blockColor!30, 
        text width=1.8cm, 
        text centered, 
        rounded corners=0.2cm, 
        minimum height=1.8em,
        font=\sffamily\scriptsize\bfseries,
        drop shadow={shadow xshift=0.3mm, shadow yshift=-0.3mm, opacity=0.3}
    },
    smallblock/.style={
        rectangle, 
        draw=black!50, 
        fill=smallBlockColor!30, 
        text width=1.8cm, 
        text centered, 
        rounded corners=0.15cm, 
        minimum height=1.4em,
        font=\sffamily\scriptsize\bfseries,
        drop shadow={shadow xshift=0.2mm, shadow yshift=-0.2mm, opacity=0.3}
    },
    circ/.style={
        circle, 
        draw=black!50, 
        fill=white,
        minimum size=0.6cm,
        font=\sffamily\scriptsize\bfseries,
        drop shadow={shadow xshift=0.1mm, shadow yshift=-0.1mm, opacity=0.3}
    },
    line/.style={
        draw=black!70,
        -latex',
        line width=0.8pt,
        shorten >=1pt,
        shorten <=1pt
    },
    skip line/.style={
        draw=red!70,
        -latex',
        dashed,
        line width=0.8pt,
        shorten >=2pt,
        shorten <=2pt
    },
]
    % Attention-based Architecture (content unchanged, but with adjusted spacing)
    \node[block, fill=inputColor!30] (input) at (0,0) {Input};
    \node[circ] (add1) [below=0.7cm of input] {+};
    \node[smallblock] (pos_enc) [left=1.5cm of add1] {Positional Embedding};
    
    \node[block] (rn1) [below=0.7cm of add1] {RMSNorm};
    \node[block] (attn) [below=0.4cm of rn1] {Attention};
    \node[circ] (add2) [below=0.4cm of attn] {+};
    \node[block] (rn2) [below=0.4cm of add2] {RMSNorm};
    \node[block] (ffn) [below=0.4cm of rn2] {MoE*};
    \node[circ] (add3) [below=0.4cm of ffn] {+};
    \node[fit=(rn1) (add3), inner sep=0.2cm, draw=transformerColor, thick, rounded corners=0.3cm] (block) {};
    
    \node[block, fill=outputColor!30] (output) [below=0.7cm of block] {Output};
    
    % Connections
    \path [line] (input) -- (add1);
    \path [line] (pos_enc) -- (add1);
    \path [line] (add1) -- (rn1);
    \path [line] (rn1) -- (attn);
    \path [line] (attn) -- (add2);
    \path [line] (add2) -- (rn2);
    \path [line] (rn2) -- (ffn);
    \path [line] (ffn) -- (add3);
    \path [line] (add3) -- (output);
    
    % Skip Connections
    \path [skip line] (add1) -| ($(block.east)+(0,1.0)$) -- (add2);
    \path [skip line] (add2) -| ($(block.west)-(0,0.65)$) |- (add3);
    
    \node[right=0.05cm of block, font=\sffamily\scriptsize\bfseries, transformerColor] (nx) {$\times N$};
    
    \node[fit=(input) (output) (block) (nx) (pos_enc), inner sep=0.3cm, draw=borderColor, line width=1pt, rounded corners=0.5cm] (border) {};

    \node[anchor=south west, font=\scriptsize] at ($(border.south west) + (0.2, 0.2)$) {*Gated Experts};
\end{tikzpicture}
\vspace{0.2em}
(b) Attention-based architecture.
\end{minipage}
\hfill
\begin{minipage}[t]{0.3\textwidth}
\centering
\begin{tikzpicture}[scale=0.5,
    % Define styles
    block/.style={
        rectangle, 
        draw=black!50, 
        fill=blockColor!30, 
        text width=1.8cm, 
        text centered, 
        rounded corners=0.2cm, 
        minimum height=1.8em,
        font=\sffamily\scriptsize\bfseries,
        drop shadow={shadow xshift=0.3mm, shadow yshift=-0.3mm, opacity=0.3}
    },
    smallblock/.style={
        rectangle, 
        draw=black!50, 
        fill=smallBlockColor!30, 
        text width=1.8cm, 
        text centered, 
        rounded corners=0.15cm, 
        minimum height=1.4em,
        font=\sffamily\scriptsize\bfseries,
        drop shadow={shadow xshift=0.2mm, shadow yshift=-0.2mm, opacity=0.3}
    },
    circ/.style={
        circle, 
        draw=black!50, 
        fill=white,
        minimum size=0.6cm,
        font=\sffamily\scriptsize\bfseries,
        drop shadow={shadow xshift=0.1mm, shadow yshift=-0.1mm, opacity=0.3}
    },
    line/.style={
        draw=black!70,
        -latex',
        line width=0.8pt,
        shorten >=1pt,
        shorten <=1pt
    },
    skip line/.style={
        draw=red!70,
        -latex',
        dashed,
        line width=0.8pt,
        shorten >=2pt,
        shorten <=2pt
    },
]
    % Mamba-based Architecture (content unchanged, but with adjusted spacing)
    \node[block, fill=inputColor!30] (input) at (0,0) {Input};
    
    \node[block] (rn) [below=0.7cm of input] {RMSNorm};
    \node[block] (mamba) [below=0.4cm of rn] {Mamba-2};
    \node[circ] (add1) [below=0.4cm of mamba] {+};
    \node[block] (rn2) [below=0.4cm of add1] {RMSNorm};
    \node[block] (moe) [below=0.4cm of rn2] {MoE*};
    \node[circ] (add2) [below=0.4cm of moe] {+};
    \node[fit=(rn) (add2), inner sep=0.2cm, draw=transformerColor, thick, rounded corners=0.3cm] (block) {};
    
    \node[block, fill=outputColor!30] (output) [below=0.7cm of block] {Output};
    
    % Connections
    \path [line] (input) -- (rn);
    \path [line] (rn) -- (mamba);
    \path [line] (mamba) -- (add1);
    \path [line] (add1) -- (rn2);
    \path [line] (rn2) -- (moe);
    \path [line] (moe) -- (add2);
    \path [line] (add2) -- (output);
    
    % Skip Connections
    \path [skip line] (input) -| ($(block.east)+(0,0.9)$) |- (add1);
    \path [skip line] (add1) -| ($(block.west)-(0,0.65)$) |- (add2);
    
    \node[right=0.05cm of block, font=\sffamily\scriptsize\bfseries, transformerColor] (nx) {$\times N$};
    
    \node[fit=(input) (output) (block) (nx), inner sep=0.3cm, draw=borderColor, line width=1pt, rounded corners=0.5cm] (border) {};
\end{tikzpicture}
\vspace{0.2em}\\
(c) Mamba-based architecture.
\end{minipage}

\caption{Comparison of STU-based, Attention-based, and Mamba-based model architectures.}
\label{fig:model-architectures}
\end{figure}

%%%%%%%%%%%%%%%%%%%%%%%%%%%%%%%%%%%%%%%%%%%%%%%%%%%%%%%%%%%%%%%%%%%%%%%%%%

\subsection{Additional experiments with robotics data}
\label{appendix:robotics}
In this section we give the remaining experimental results over robotics data for the other two MuJoCo tasks: HalfCheetah-v1 and Walker2D-v1. Compared to Ant-v1 (shown in Section 2 in the main paper), these two tasks only involve motion in a 2D plane. As a result, the state representations for HalfCheetah-v1 and Walker2D-v1 are inherently less complicated, potentially affecting the relative performance of different models on these tasks in theory.

\begin{figure}[H]
    \centering
    \begin{minipage}{0.4\textwidth}
        \centering
        \includegraphics[width=\textwidth]{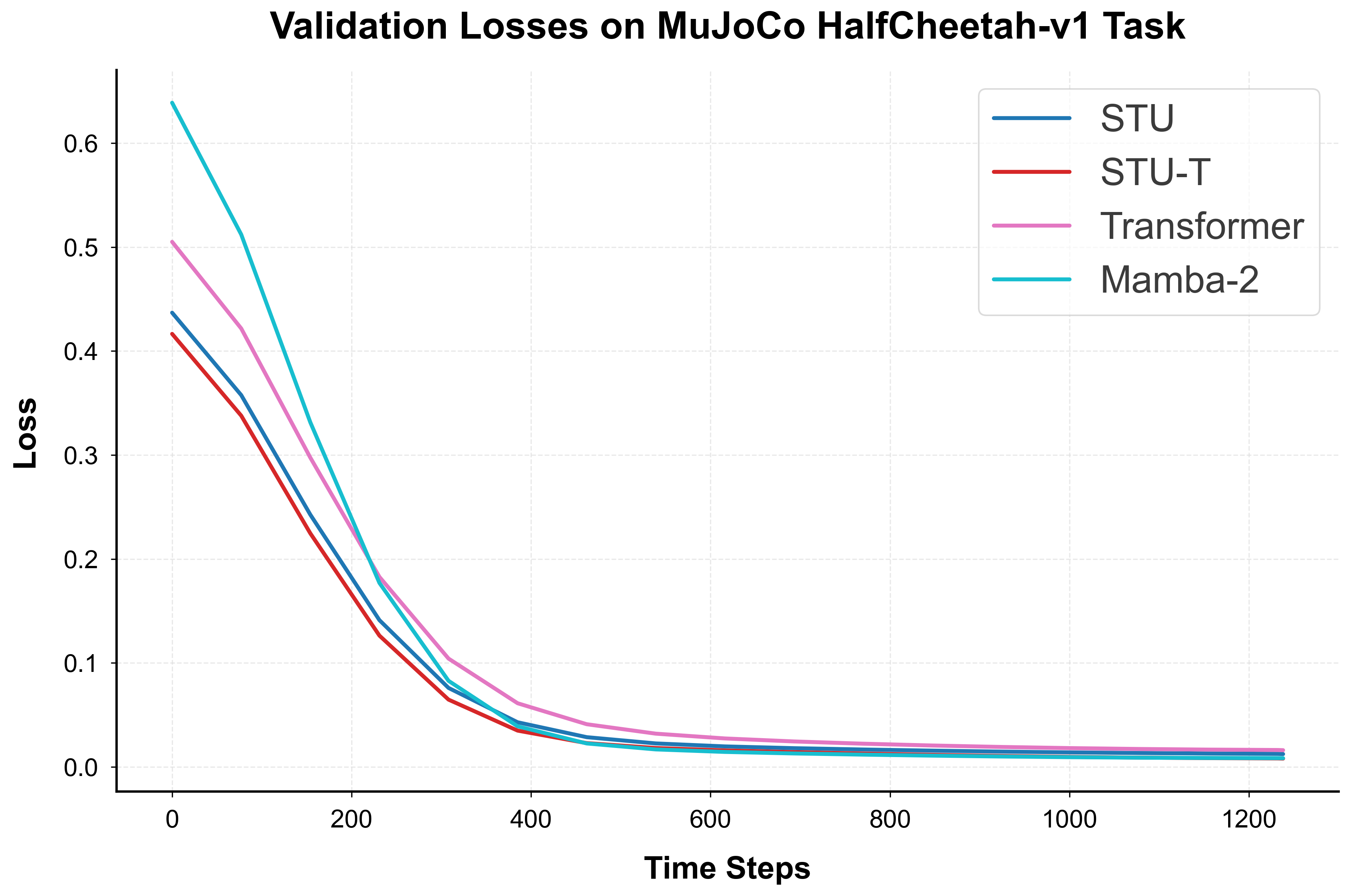}
        \caption{\texttt{HalfCheetah-v1} comparative training results\label{fig:val-losses-HalfCheetah}.}
    \end{minipage}%
    \hspace{0.04\textwidth}
    \begin{minipage}{0.4\textwidth}
        \centering
        \captionof{table}{\texttt{HalfCheetah-v1} comparative validation loss results\label{table:val-losses-HalfCheetah}.}
        \begin{tabular}{lcc}
          \toprule
          Model & Validation Loss \\
          \midrule
          STU & 0.0119 \\
          STU-T & \textbf{0.0076} \\
          Transformer & 0.0157 \\
          Mamba-2 & 0.0081 \\
          \bottomrule
        \end{tabular}
    \end{minipage}
\end{figure}

\begin{figure}[H]
    \centering
    \begin{minipage}{0.4\textwidth}
        \centering
        \includegraphics[width=\textwidth]{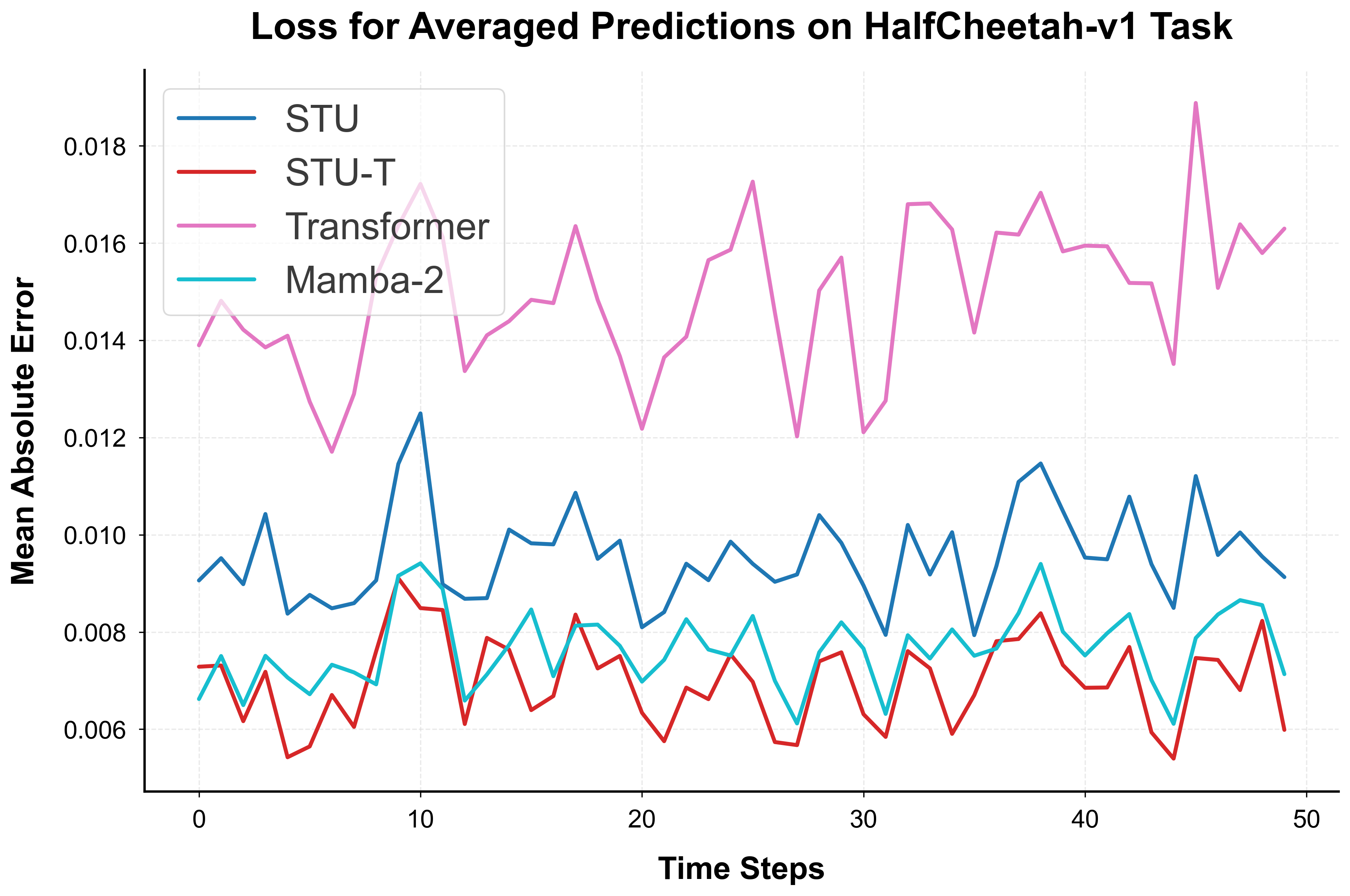}
        \caption{\texttt{HalfCheetah-v1} comparative next-step prediction results (with losses averaged over 500 predictions for each model)\label{fig:pred-losses-HalfCheetah}.}
    \end{minipage}%
    \hspace{0.04\textwidth}
    \begin{minipage}{0.4\textwidth}
        \centering
        \includegraphics[width=\textwidth]{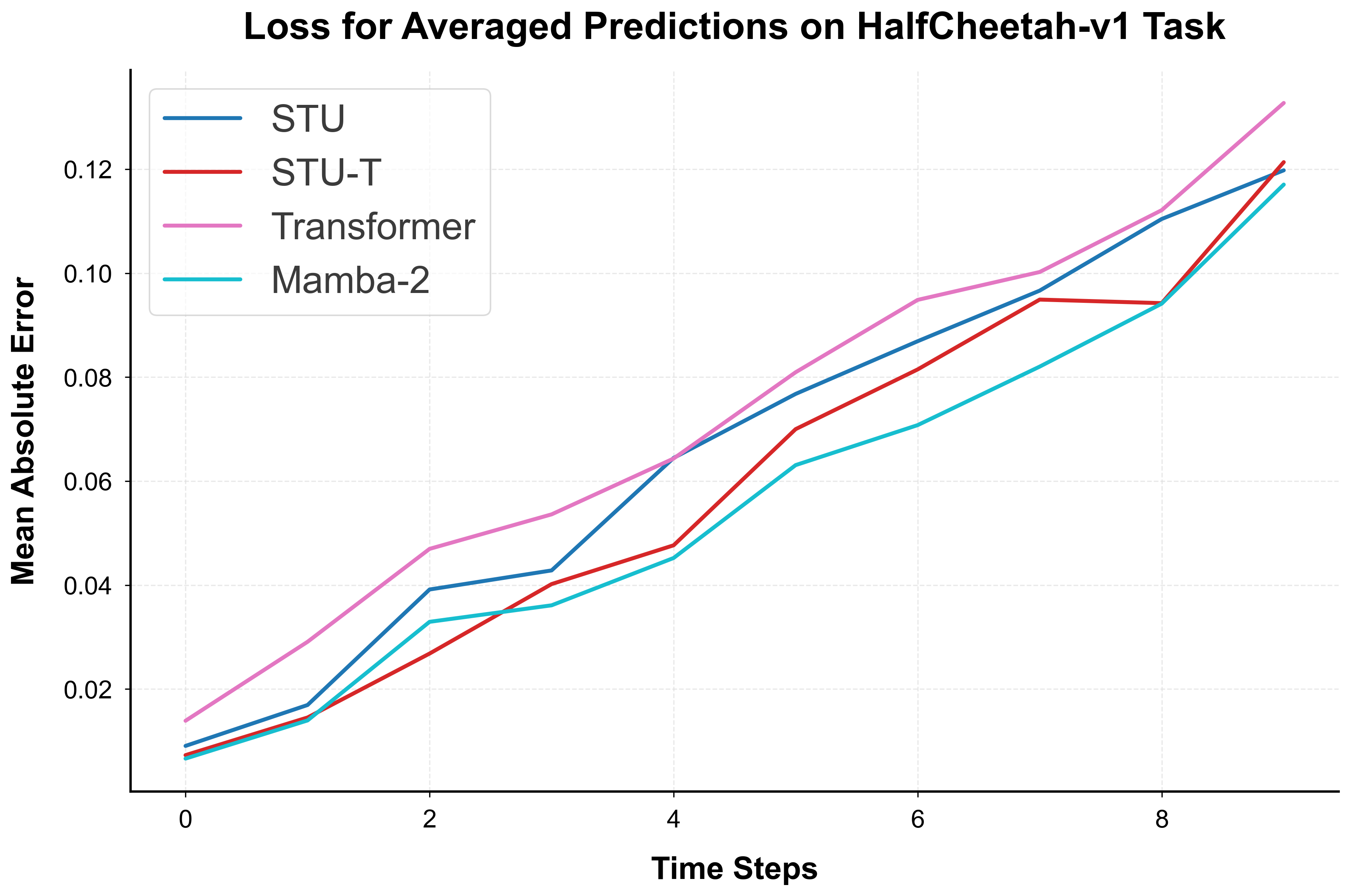}
        \caption{\texttt{HalfCheetah-v1} comparative auto-regressive next-step prediction results (with losses averaged over 500 predictions for each model)\label{fig:pred-losses-HalfCheetah-ar}.}
    \end{minipage}
\end{figure}

\begin{figure}[H]
    \centering
    \begin{minipage}{0.4\textwidth}
        \centering
        \includegraphics[width=\textwidth]{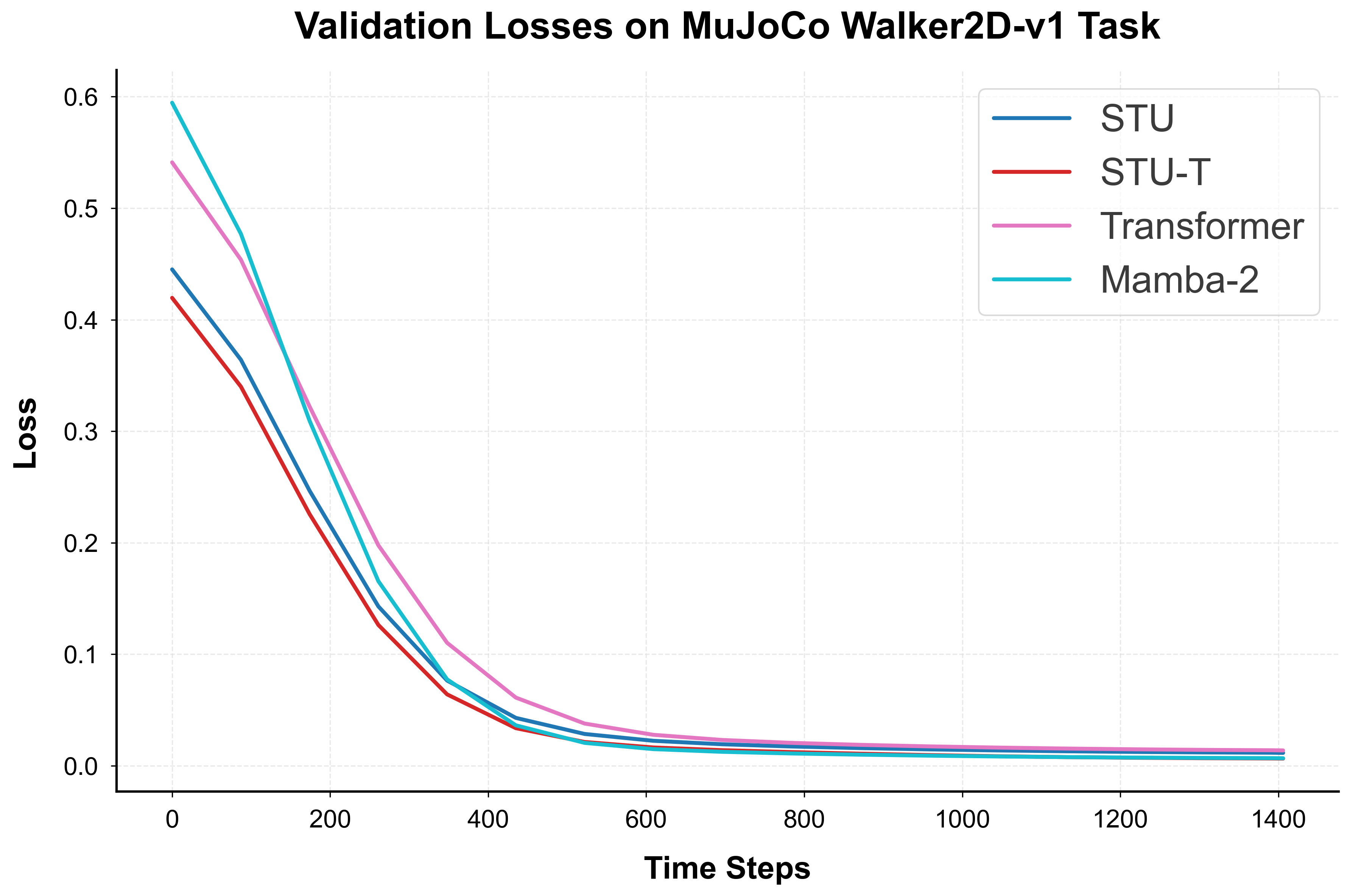}
        \caption{\texttt{Walker2D-v1} comparative training results\label{fig:val-losses-Walker2D}.}
    \end{minipage}%
    \hspace{0.04\textwidth}
    \begin{minipage}{0.4\textwidth}
        \centering
        \captionof{table}{\texttt{Walker2D-v1} comparative validation loss results\label{table:val-losses-Walker2D}.}
        \begin{tabular}{lcc}
          \toprule
          Model & Validation Loss \\
          \midrule
          STU & 0.0112 \\
          STU-T & \textbf{0.0062} \\
          Transformer & 0.0134 \\
          Mamba-2 & 0.0066 \\
          \bottomrule
        \end{tabular}
    \end{minipage}
\end{figure}

\begin{figure}[H]
    \centering
    \begin{minipage}{0.4\textwidth}
        \centering
        \includegraphics[width=\textwidth]{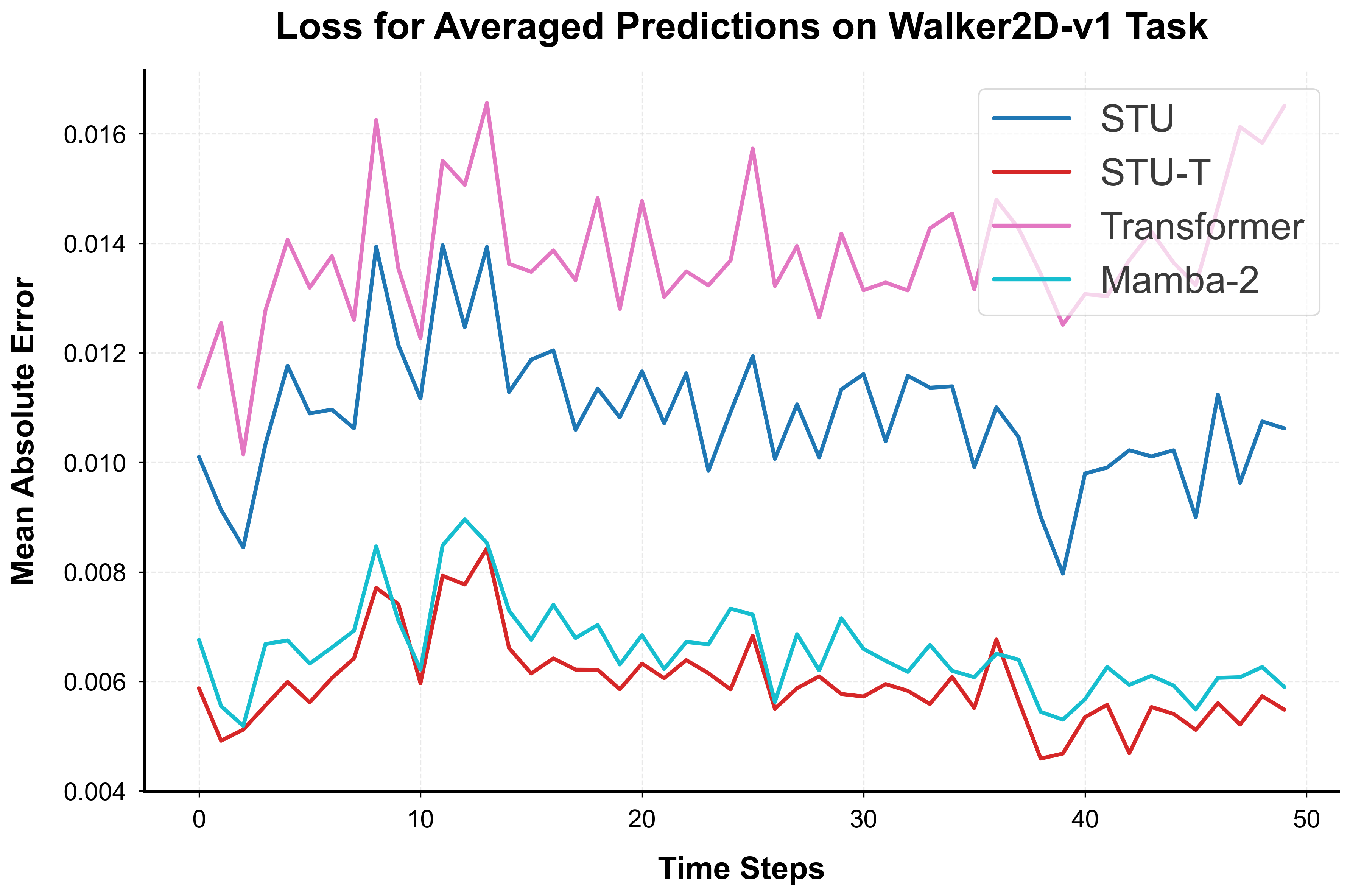}
        \caption{\texttt{Walker2D-v1} comparative next-step prediction results (with losses averaged over 500 predictions for each model).}
        \label{fig:pred-losses-Walker2D}
    \end{minipage}%
    \hspace{0.04\textwidth}
    \begin{minipage}{0.4\textwidth}
        \centering
        \includegraphics[width=\textwidth]{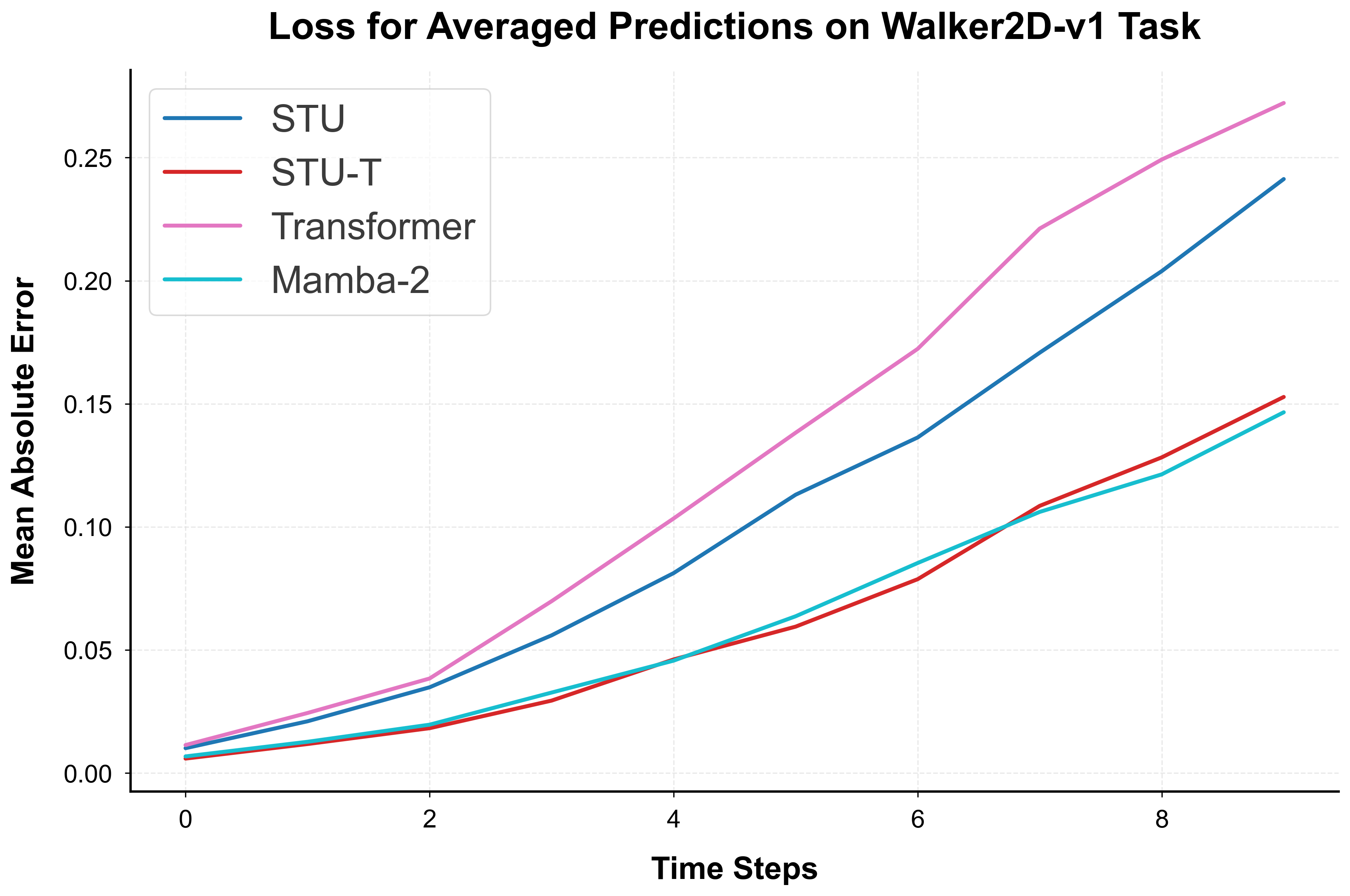}
        \caption{\texttt{Walker2D-v1} comparative auto-regressive next-step prediction results (with losses averaged over 500 predictions for each model).}
        \label{fig:pred-losses-Walker2D-ar}
    \end{minipage}
\end{figure}

In this setting, the comparison of the models' performances still remains consistent, with Transformer showing the least competitive results, Mamba-2 demonstrating significant improvements over Transformer, and STU-T outperforming both Transformer and Mamba-2. HalfCheetah-v1 task's training results are shown in Figure \ref{fig:val-losses-HalfCheetah} and Table \ref{table:val-losses-HalfCheetah}, (auto-regressive) next-step predictions results in Figures \ref{fig:pred-losses-HalfCheetah} and \ref{fig:pred-losses-HalfCheetah-ar}. Walker2D-v1 task's training results are shown in Figure \ref{fig:val-losses-Walker2D} and Table \ref{table:val-losses-Walker2D}, (auto-regressive) next-step predictions results in Figures \ref{fig:pred-losses-Walker2D} and \ref{fig:pred-losses-Walker2D-ar}.

\begin{figure}[H]
    \centering
    \begin{minipage}{0.48\textwidth}
        \begin{figure}[H]
            \centering
            \includegraphics[width=\textwidth]{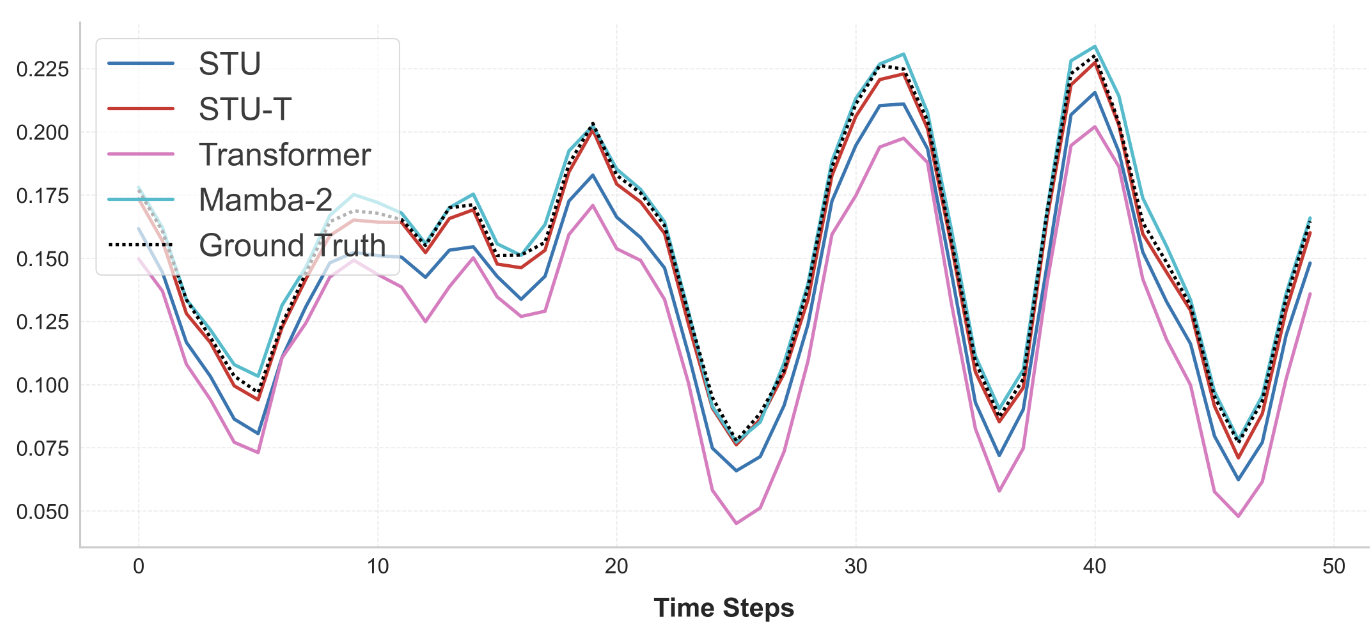}
            \caption{\texttt{Ant-v1} next-step prediction (averaged) trajectory of the angle between the ant's two links on the back right.}
            \label{pred_traj_1}
        \end{figure}
    \end{minipage}
    \hfill
    \begin{minipage}{0.48\textwidth}
        \begin{figure}[H]
            \centering
            \includegraphics[width=\textwidth]{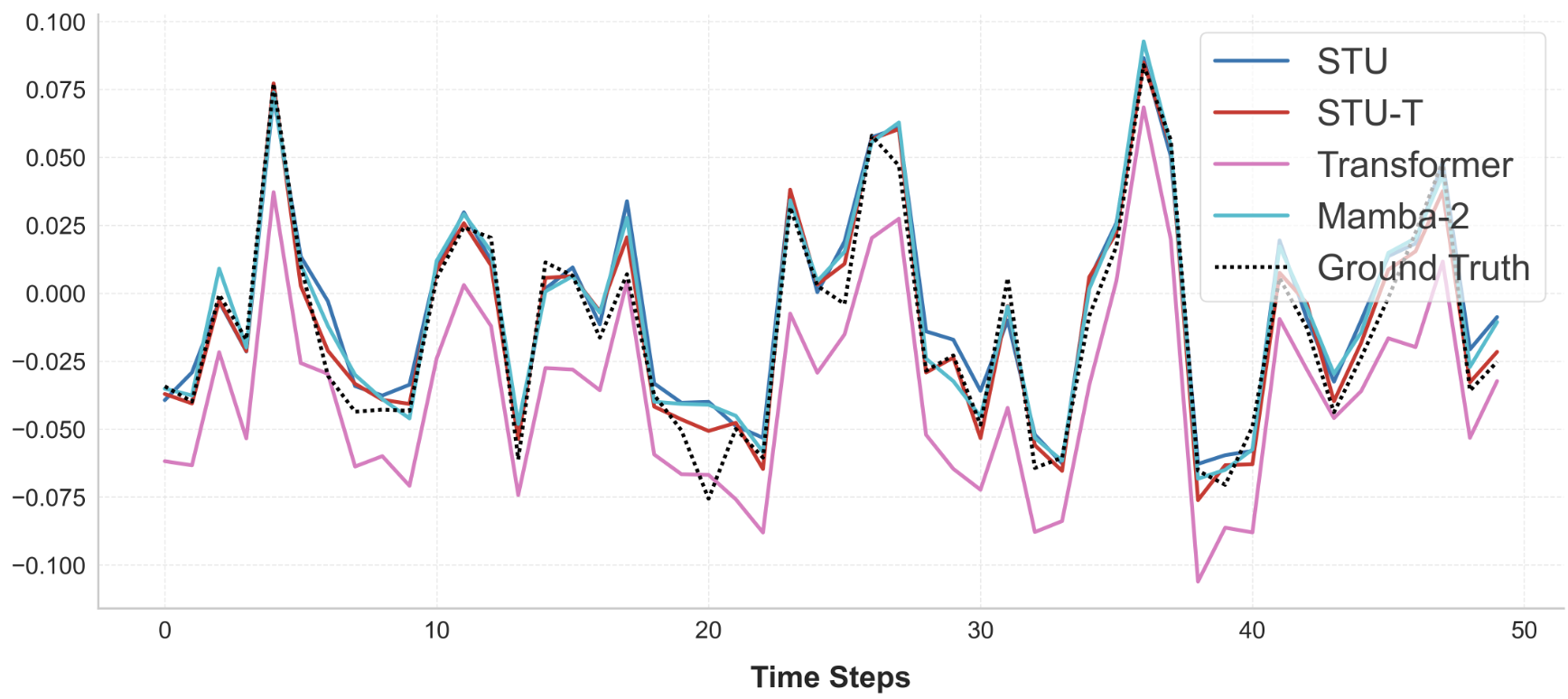}
            \caption{\texttt{Ant-v1} next-step prediction (averaged) trajectory of the x-coordinate angular velocity of the torso.}
            \label{pred_traj_2}
        \end{figure}
    \end{minipage}
\end{figure}

\newpage
For the (auto-regressive) next-step prediction figures presented above and in Section \ref{robotics-experiments} of the paper, we mainly focus on showing the loss, i.e. the difference between the models' predictions and the ground truths, averaged over 500 predictions. To further demonstrate what these predictions actually look like, below we give two prediction trajectories plotted across 50 time steps specifically on two example features of \texttt{Ant-v1}: the angle between the two links on the back right (Figure \ref{pred_traj_1}) and the x-coordinate angular velocity of the torso (Figure \ref{pred_traj_2}).

\vspace{-1em}
\vspace{0.25cm}

\subsubsection{Hyperparameters for robotics experiments}
\label{robotics-hyperparameters}
We conducted comprehensive ablation studies to investigate the impact of various hyperparameters on the performance of STU, STU-T, and Transformer models in the context of robotics tasks. These studies explore the effects of model width, depth, and input noise on the final Mean Squared Error (MSE) loss. Tables \ref{table:axes-stu}, \ref{table:axes-stu-t}, and \ref{table:axes-transformer} present the results of these experiments.

\begin{table}[ht]
  \caption{Ablation studies for STU models.}
  \label{table:axes-stu}
  \centering
  \small
  \begin{tabular}{lccccc}
    \toprule
    Model & Parameter Count & Width & Layers & Noise/Frequency & MSE Loss \\
    \midrule
    STU & 0.18M & \textbf{32} & 4 & 0.0/0.0 & 0.0217 \\
    STU & 0.63M & \textbf{64} & 4 & 0.0/0.0 & 0.0139 \\
    STU & 1.33M & \textbf{96} & 4 & 0.0/0.0 & 0.0120 \\
    STU & 2.30M & \textbf{128} & 4 & 0.0/0.0 & \textbf{0.0108} \\
    \midrule
    STU & 0.05M & 32 & \textbf{1} & 0.0/0.0 & 0.0449 \\
    STU & 0.09M & 32 & \textbf{2} & 0.0/0.0 & 0.0306 \\
    STU & 0.18M & 32 & \textbf{4} & 0.0/0.0 & 0.0217 \\
    STU & 0.27M & 32 & \textbf{6} & 0.0/0.0 & \textbf{0.0203} \\
    \midrule
    STU & 0.18M & 32 & 4 & \textbf{0.0/0.0} & \textbf{0.0217} \\
    STU & 0.18M & 32 & 4 & \textbf{0.1/0.1} & 0.0357 \\
    STU & 0.18M & 32 & 4 & \textbf{0.5/0.1} & 0.0561 \\
    \bottomrule
  \end{tabular}
\end{table}

\begin{table}[ht]
  \caption{Ablation studies for STU-T models.}
  \label{table:axes-stu-t}
  \centering
  \small
  \begin{tabular}{lccccc}
    \toprule
    Model & Parameter Count & Width & Layers & Noise/Frequency & MSE Loss \\
    \midrule
    STU-T & 0.06M & \textbf{32} & 4 & 0.0/0.0 & 0.0239 \\
    STU-T & 0.12M & \textbf{64} & 4 & 0.0/0.0 & 0.0146 \\
    STU-T & 0.20M & \textbf{96} & 4 & 0.0/0.0 & 0.0116 \\
    STU-T & 0.28M & \textbf{128} & 4 & 0.0/0.0 & \textbf{0.0105} \\
    \midrule
    STU-T & 0.02M & 32 & \textbf{1} & 0.0/0.0 & 0.0464 \\
    STU-T & 0.03M & 32 & \textbf{2} & 0.0/0.0 & 0.0328 \\
    STU-T & 0.06M & 32 & \textbf{4} & 0.0/0.0 & 0.0239 \\
    STU-T & 0.09M & 32 & \textbf{6} & 0.0/0.0 & \textbf{0.0218} \\
    \midrule
    STU-T & 0.06M & 32 & 4 & \textbf{0.0/0.0} & \textbf{0.0239} \\
    STU-T & 0.06M & 32 & 4 & \textbf{0.1/0.1} & 0.0429 \\
    STU-T & 0.06M & 32 & 4 & \textbf{0.5/0.1} & 0.0688 \\
    \bottomrule
  \end{tabular}
\end{table}

\begin{table}[H]
  \caption{Ablation studies for Transformer models.}
  \label{table:axes-transformer}
  \centering
  \small
  \begin{tabular}{lccccc}
    \toprule
    Model & Parameter Count & Width & Layers & Noise/Frequency & MSE Loss \\
    \midrule
    Transformer & 0.07M & \textbf{32} & 4 & 0.0/0.0 & 0.0472 \\
    Transformer & 0.17M & \textbf{64} & 4 & 0.0/0.0 & 0.0294 \\
    Transformer & 0.30M & \textbf{96} & 4 & 0.0/0.0 & 0.0214 \\
    Transformer & 0.47M & \textbf{128} & 4 & 0.0/0.0 & \textbf{0.0204} \\
    \midrule
    Transformer & 0.02M & 32 & \textbf{1} & 0.0/0.0 & 0.0545 \\
    Transformer & 0.04M & 32 & \textbf{2} & 0.0/0.0 & 0.0464 \\
    Transformer & 0.07M & 32 & \textbf{4} & 0.0/0.0 & 0.0472 \\
    Transformer & 0.10M & 32 & \textbf{6} & 0.0/0.0 & \textbf{0.0462} \\
    \midrule
    Transformer & 0.07M & 32 & 4 & \textbf{0.0/0.0} & \textbf{0.0472} \\
    Transformer & 0.07M & 32 & 4 & \textbf{0.1/0.1} & 0.0637 \\
    Transformer & 0.07M & 32 & 4 & \textbf{0.5/0.1} & 0.0961 \\
    \bottomrule
  \end{tabular}
\end{table}

Note that the noise level refers to the standard deviation of Gaussian noise added to the input data, while the frequency represents the probability of applying this noise to each data point. For example, $0.1/0.1$ means that, on average, Gaussian noise with a standard deviation of 0.1 is applied to 10\% of the data points.

\subsubsection{Hessian heat maps for robotics experiments}

\begin{figure}[H]
\centering
\begin{minipage}[t]{0.32\textwidth}
    \centering
    \includegraphics[width=\textwidth]{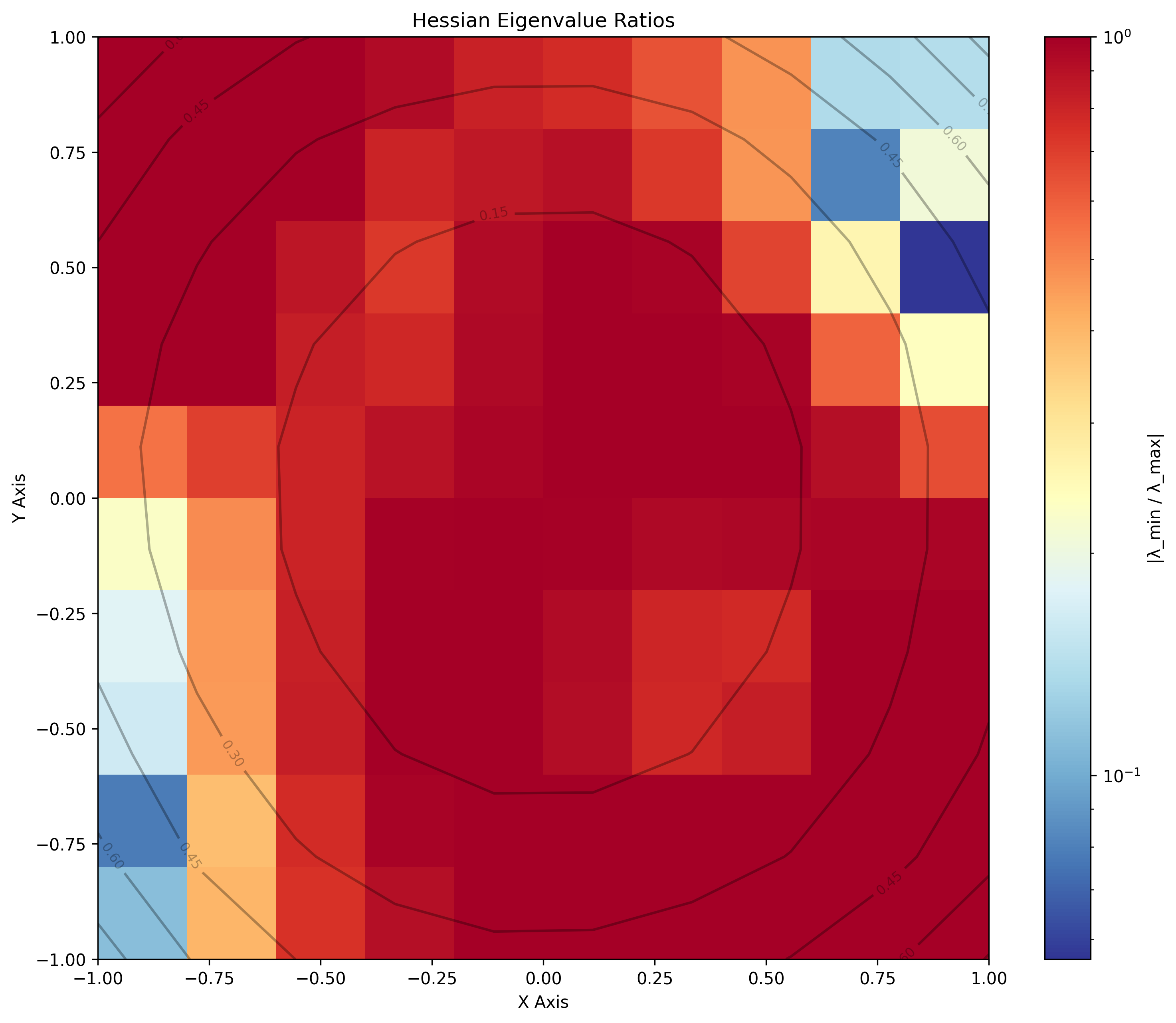}
    (a) Transformer model.
\end{minipage}
\hfill
\begin{minipage}[t]{0.32\textwidth}
    \centering
    \includegraphics[width=\textwidth]{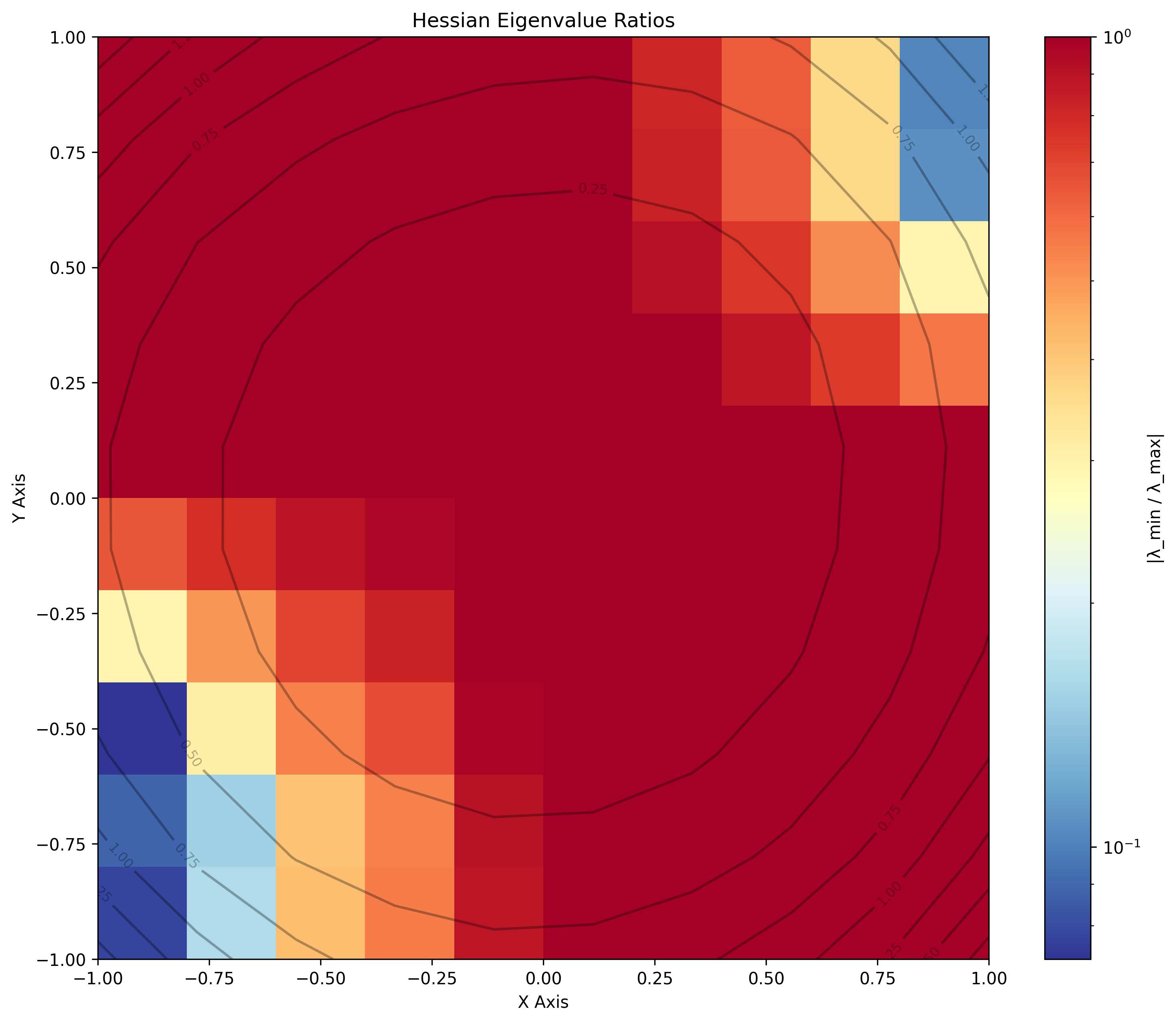}
    (b) STU model.
\end{minipage}
\hfill
\begin{minipage}[t]{0.32\textwidth}
    \centering
    \includegraphics[width=\textwidth]{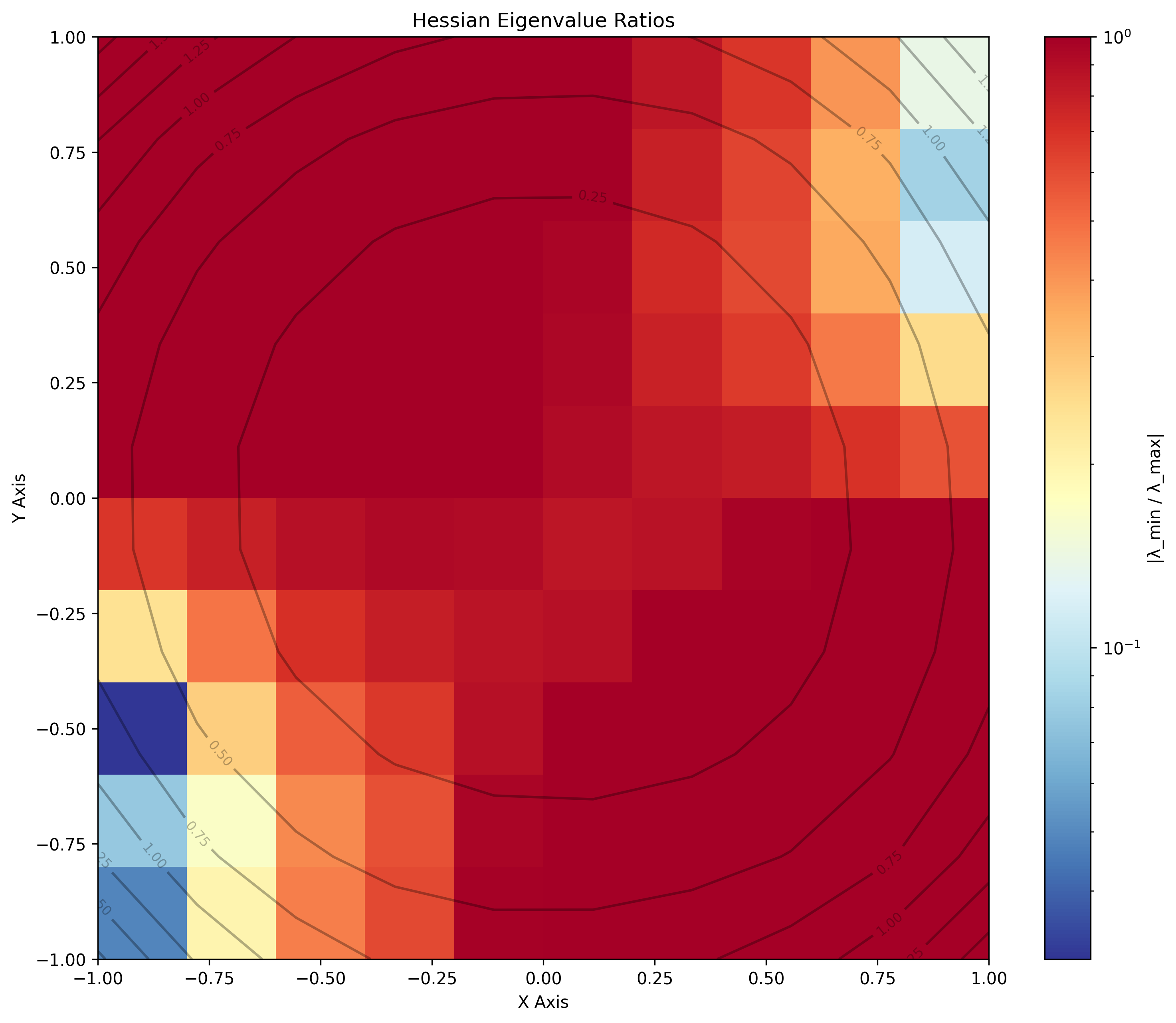}
    (c) STU-T model.
\end{minipage}
\caption{Hessian heat maps for models with 4 layers and model width of 128.}
\label{fig:Models-LL}
\end{figure}

The Hessian eigenvalue ratio heat maps in Figure \ref{fig:Models-LL} support our findings on model performance in continuous control tasks. In particiular, the sharpness across the landscapes is a salient point for understanding generalization. Flatter minima are generally associated with better generalization~\cite{zhou2021theoreticallyunderstandingsgdgeneralizes}, while sharper minima can signal sensitivity to parameter changes. All models show a central red region indicating flat loss landscapes near the optimum, suggesting good generalization in this relatively simple setting. However, the STU and STU-T models exhibit smoother transitions to higher curvature regions, implying better optimization stability. In contrast, the Transformer model displays asymmetry with larger high-curvature areas, which may explain its less favorable performance. The STU-T model’s minimal high-curvature regions suggest a more globally smooth landscape, contributing to its robust performance across tasks and hyperparameter settings.

\label{loss-landscapes}

%% file: appendix_llm.tex
%%%%%%%%%%%%%%%%%%%%%%%%%%%%%%%%%%%%%%%%%%%%%%%%%%%%%%%%%%%%%%%%%%%%%%%%%%

\subsection{LLM experiments}
\label{appendix:model-configs}
\begin{table}[H]
  \caption{Model and training configuration details for 2B LLM training run.}
  \label{table:model-configs-2B}
  \centering
  \scriptsize
  \begin{tabular}{llcc}
    \toprule
    \multicolumn{4}{c}{\textbf{Model Architecture}} \\
    \midrule
    \text{ }            & \textbf{Description}                        & \textbf{Flash STU} & \textbf{Transformer} \\
    \midrule
    \textbf{Parameter Count}      & Total number of parameters                  & 2,672M             & 2,667M              \\
    \textbf{Embedding Dimension}  & Dimensionality of embedding space                & 1,536              & 1,536               \\
    \textbf{Number of Heads}      & Attention heads (not multi-queried or multi-grouped)                             & 8                  & 8                   \\
    \textbf{Number of Layers}     & Transformer layers                          & 26                 & 25                  \\
    \textbf{ALiBi Attention}    & Attention scores modification using linear biases          & Yes (interpolation factor: $0.25$)                & Yes (interpolation factor: $0.25$)                  \\
    \textbf{Sliding Window Size}          & Sliding window attention context lookback size               & 1,024              & 8,192               \\
    \textbf{Sequence Length (Training)}      & Input sequence length during training               & 8,192              & 8,192               \\
    \textbf{Sequence Length (Inference)}      & Input sequence length during inference via position interpolation              & 32,768              & 32,768              \\
    \textbf{Vocabulary Size}      & Size of the model's vocabulary              & 200,064            & 200,064             \\
    \textbf{MLP Expansion Factor} & Expansion factor in MLP layers              & 12                 & 12                  \\
    \textbf{Bias}                 & Use of bias terms in linear layers                         & No                 & No                  \\
    \textbf{Dropout}              & Dropout rate                                & 0.0                & 0.0                 \\
    \textbf{Number of Filters}    & Number of filters (Flash STU only)          & 24                 & --                  \\
    \textbf{Use Hankel$_L$}       & Alternative Hankel matrix (Flash STU only)       & No                 & --                  \\
    \textbf{Learnable Filters}    & Learnable filters (Flash STU only)          & Yes                & --                  \\
    \midrule
    \multicolumn{4}{c}{\textbf{Training and Optimization}} \\
    \midrule
    \textbf{Epochs}               & Number of training epochs                    & 1                  & 1                   \\
    \textbf{Global Batch Size}    & Number of tokens processed per step            & 524,288            & 524,288             \\
    \textbf{Micro Batch Size}     & Batch size per GPU                           & 1                  & 1                   \\
    \textbf{Gradient Accumulation Steps} & Number of steps before performing a gradient update & 8 & 8 \\
    \textbf{Warmup Steps}         & Number of warmup steps                       & 1,907              & 1,907               \\
    \textbf{Evaluation Period}    & Evaluation frequency (steps)                 & 25                 & 25                  \\
    \textbf{Max Grad Norm}        & Maximum gradient norm for clipping           & 1.0                & 1.0                 \\
    \midrule
    \multicolumn{4}{c}{\textbf{Optimizer Configuration}} \\
    \midrule
    \textbf{Optimizer}                & Optimizer type                            & AdamW              & AdamW               \\
    \textbf{Learning Rate Schedule}   & LR scheduling strategy                    & Linear decay with warmup & Linear decay with warmup \\
    \textbf{Max Learning Rate}        & Maximum learning rate                     & $3.0 \times 10^{-4}$ & $3.0 \times 10^{-4}$ \\
    \textbf{Min Learning Rate}        & Minimum learning rate                     & $3.0 \times 10^{-5}$ & $3.0 \times 10^{-5}$ \\
    \textbf{Betas}                    & Optimizer betas                           & (0.9, 0.999)       & (0.9, 0.999)        \\
    \textbf{Epsilon}                  & Optimizer epsilon                         & $1.0 \times 10^{-8}$ & $1.0 \times 10^{-8}$ \\
    \textbf{Weight Decay}             & Weight decay factor                       & $1.0 \times 10^{-2}$ & $1.0 \times 10^{-2}$ \\
    \textbf{AMSGrad}                  & Use AMSGrad variant                       & No                 & No                  \\
    \textbf{Fused}                    & Use fused optimizer                       & Yes                & Yes                 \\
    \midrule
    \multicolumn{4}{c}{\textbf{Optimization Techniques}} \\
    \midrule
    \textbf{Activation Checkpointing} & Enable activation checkpointing           & Yes                & Yes                 \\
    \textbf{Use Flash FFT}            & Enable Flash FFT (Flash STU only)          & Yes                & --                  \\
    \textbf{Use Tensordot Approx.}    & Enable tensordot approximation            & Yes                & --                  \\
    \textbf{Use Attention}            & Enable attention mechanism                & Yes                & Yes                 \\
    \textbf{Softcap}                  & Softcap threshold                         & 50.0               & 50.0                \\
    \textbf{Torch Compile}            & Enable Torch compile optimization         & No                 & No                  \\
    \midrule
    \multicolumn{4}{c}{\textbf{Distributed Training Configuration}} \\
    \midrule
    \textbf{FSDP}                     & Fully Sharded Data Parallel                & Yes                & Yes                 \\
    \textbf{DDP}                      & Distributed Data Parallel                  & No                 & No                  \\
    \textbf{Mixed Precision}          & Use mixed precision training               & Yes                & Yes                 \\
    \textbf{Torch Dtype}              & Data type for PyTorch tensors              & \texttt{bfloat16}  & \texttt{bfloat16}   \\
    \textbf{CPU Offload}              & Offload computations to CPU                & No                 & No                  \\
    \textbf{Sharding Strategy}        & Strategy for model sharding                & Full Shard         & Full Shard          \\
    \textbf{FSDP Modules}             & Modules to apply FSDP                      & (STU, Attention, MLP)    & (Attention, MLP)          \\
    \textbf{State Dict Type}          & Type of state dictionary                   & Full               & Full                \\
    \textbf{Auto Wrap Policy}         & Policy for automatic wrapping              & Partial            & Partial             \\
    \textbf{Backward Prefetch}        & Backward prefetch strategy                 & Backward Pre       & Backward Pre        \\
    \textbf{Forward Prefetch}         & Forward prefetch                           & No                 & No                  \\
    \textbf{Sync Module States}       & Synchronize module states across replicas  & Yes                & Yes                 \\
    \textbf{Use Original Params}      & Use original parameters during training    & Yes                & Yes                 \\
    \bottomrule
  \end{tabular}
\end{table}

\begin{table}[H]
  \caption{Model and training configuration details for 500M LLM training runs.}
  \label{table:model-configs-500M}
  \centering
  \footnotesize
  \begin{tabular}{lcccc}
    \toprule
    & \textbf{Flash STU} & \textbf{Transformer} & \textbf{Mamba-2} & \textbf{Mamba-2 Hybrid} \\
    \midrule
    \multicolumn{5}{l}{\textbf{Model Architecture}} \\
    \midrule
    Parameters & 550M & 564M & 561M & 546M \\
    Hidden Dim & 896 & 896 & 1024 & 896 \\
    Attention Heads & 8 & 8 & 32 & 32 \\
    Layers & 12 & 12 & 54 & 56 \\
    Sequence Length & 8192 & 8192 & 8192 & 8192 \\
    Attention Context Window & 1024 & 8192 & -- & 1024 \\
    Tied Weights & True & True & True & True \\
    SSM State Dim & -- & -- & 128 & 128 \\
    MLP Scale & 12 & 12 & 2 & 2 \\
    Softcap & 50 & 50 & -- & 50 \\
    RoPE Theta & 10,000 & 10,000 & -- & -- \\
    ALiBi & False & False & -- & True \\
    \midrule
    \multicolumn{5}{l}{\textbf{Training}} \\
    \midrule
    Global Batch Size & 524{,}288 & 524{,}288 & 524{,}288 & 524{,}288 \\
    Min/Max Learning Rate & 3e-4/3e-5 & 3e-4/3e-5 & 3e-4/3e-5 & 3e-4/3e-5 \\
    LR Schedule & \multicolumn{4}{c}{Linear decay w/ warmup} \\
    Training Precision & \multicolumn{4}{c}{\texttt{bfloat16}} \\
    \bottomrule
  \end{tabular}
\end{table}

% \begin{figure}[H]
%     \centering
%     \includegraphics[width=\textwidth]{figures/flash_stu_learnable_vs_transformer_train.png}
%     \caption{Comparison of Flash STU and Transformer Training Performance.}
%     \label{fig:flash_stu_learnable_vs_transformer_train}
% \end{figure}

% \begin{figure}[H]
%     \centering
%     \includegraphics[width=0.7\textwidth]{figures/all_models.png}
%     \caption{Comparison of STU-T, Flash STU, and Transformer on validation set.}
%     \label{fig:all-models}
% \end{figure}

\label{appendix:flash-stu-output}

To further illustrate the capabilities of the Flash STU 2B model, we present a sample output generated using a temperature of $1.0$ and \texttt{top\_k} sampling with $k=50$:

\vspace{0.15in}
\begin{displayquote}
\textbf{Starter prompt:} \textit{"In a shocking finding, scientist discovered a herd of unicorns living in a remote, previously unexplored valley, in the Andes Mountains. Even more surprising to the researchers was the fact that the unicorns spoke perfect English."}

\textbf{Generated output:} \textit{"In a shocking finding, scientist discovered a herd of unicorns living in a remote, previously unexplored valley, in the Andes Mountains. Even more surprising to the researchers was the fact that the unicorns spoke perfect English. Researchers now believe that these animals could now be a kind of hominid species.} \\
\\ \textit{This discovery was recently made by scientists from the University of London. Previously, scientists had thought that this species stood only 5 feet (1.2 meters) for 15 minutes. But this new discovery was the first time a unicorn's eyes were ever revealed.} \\
\\ \textit{It was a perfect choice to do a lot of research on the species. In the end, to this day, we still can't say why the unicorns lived in this remote world - they were just another big black monkey. But the discovery proved that these unicorns used words and were able to communicate with other beings using the same words.} \\
\\ \textit{The study, which involved researchers from both Washington State University and Cornell University, confirmed that the unicorn family has existed for more than 10,000 years. It was also the first the species was studied and studied in its entirety.} \\
\\ \textit{'While there are also other rare wild animals like unicorns and hippopotami, most of these findings were not reported, but confirmed by numerous studies. In order to find their way out of an endangered species and find them, many endangered species require conservation programmes,' said lead researcher, Dr Thomas W. Huxley.} \\
\\ \textit{One of the key findings of the study in this week's issue is that the unicorns were able to walk forward and backward across the entire landscape. Their front limbs were able to reach all the way across the same landmass and in the same direction as their front legs were able to. These unique back limbs helped the animals to survive and thrive.} \\
\\ \textit{Other research has suggested that these unicorns have also been observed in the Andes Mountains. This is where the researchers believe they may have been able to find their way to the Andes. The area they found is the Amazon Valley.} \\
\\ \textit{The scientists believe that they came across an area that is part of the Amazon basin. In order to find out the true nature of this area, the researchers used satellite imagery from NASA for their research.} \\
\\ \textit{The scientists also believe that the unicorns might now also be found in other parts of the world, including North and South America. These discoveries might help scientists study an unknown area, such as the Amazonia and the Andes."}
\end{displayquote}

%%%%%%%%%%%%%%%%%%%%%%%%%%%%%%%%%%%%%%%%%%%%%%%%%%%%%%%%%%%%%%%%%%%%%%%%%%